\documentclass[sigconf]{acmart}
\renewcommand\footnotetextcopyrightpermission[1]{} 
\AtBeginDocument{%
  \providecommand\BibTeX{{%
    \normalfont B\kern-0.5em{\scshape i\kern-0.25em b}\kern-0.8em\TeX}}}

\usepackage{tabularx}
\usepackage{array,multirow}

\newcolumntype{Y}{>{\raggedright\arraybackslash}X}

\usepackage{subfigure}
\usepackage{flexisym}
\usepackage{textcomp}
\usepackage{xcolor}
\usepackage{geometry}
\usepackage{makecell}
\usepackage{color,soul}
\usepackage{textcomp}
\usepackage{color, colortbl}
\usepackage{wrapfig}
\usepackage{lscape}
\usepackage{rotating} 
\usepackage{tablefootnote}
\usepackage{amsmath}
\usepackage{caption}
\usepackage{tabularx}
\usepackage[utf8]{inputenc}
\usepackage{amsfonts}
\usepackage{caption}
\usepackage{comment}
\usepackage{makecell}
\usepackage{tabu, booktabs}

\begin{document}

\title{Identifiability of Causal-based Fairness Notions: \\ A State of the Art}

\author{Karima Makhlouf}
\affiliation{%
  \institution{Inria, École Polytechnique, IPP}
  \city{Paris}
  \country{France}
}
\email{makhlouf@lix.polytechnique.fr}

\author{Sami Zhioua}
\affiliation{%
  \institution{Inria, École Polytechnique, IPP}
  \city{Paris}
  \country{France}
}
\email{zhioua@lix.polytechnique.fr}

\author{Catuscia Palamidessi}
\affiliation{%
  \institution{Inria, École Polytechnique, IPP}
  \city{Paris}
  \country{France}
}
\email{catuscia@lix.polytechnique.fr}

\renewcommand{\shortauthors}{Makhlouf, et al.}

\begin{abstract}
Machine learning algorithms can produce biased outcome/prediction, typically, against minorities and under-represented sub-populations. Therefore, fairness is emerging as an important requirement for the large scale application of machine learning based technologies. The most commonly used fairness notions (e.g. statistical parity, equalized odds, predictive parity, etc.) are observational and rely on mere correlation between variables. These notions fail to identify bias in case of statistical anomalies such as Simpson's or Berkson's paradoxes. Causality-based fairness notions (e.g. counterfactual fairness, no-proxy discrimination, etc.) are immune to such anomalies and hence more reliable to assess fairness. The problem of causality-based fairness notions, however, is that they are defined in terms of quantities (e.g. causal, counterfactual, and path-specific effects) that are not always measurable. This is known as the identifiability problem and is the topic of a large body of work in the causal inference literature. This paper is a compilation of the major identifiability results which are of particular relevance for machine learning fairness. The results are illustrated using a large number of examples and causal graphs. The paper would be of particular interest to fairness researchers, practitioners, and policy makers who are considering the use of causality-based fairness notions as it summarizes and illustrates the major identifiability results.

\end{abstract}


\maketitle
\pagestyle{plain}
\vspace{-2mm}
\section{Introduction}
\label{intro}

Machine learning is being used to inform decisions with critical consequences on human lives such as job hiring, college admission, loan granting, and criminal risk assessment. Unfortunately, these automated decision systems have been found to consistently discriminate against certain individuals or sub-populations, typically minorities. Because the discrimination is very often unintentional, discovering and addressing it is a challenging task. The most commonly used fairness notions are observational and rely on mere correlation between variables. For example, statistical parity~\cite{dwork2012fairness} requires that the proportion of positive outcome (e.g. granting loans) is the same for all sub-populations (e.g. male and female groups).  Equal opportunity~\cite{hardt2016equality} requires that the true positive rate (TPR) is the same for all sub-populations. The main problem of correlation-based fairness notions is that they fail to detect discrimination in presence of statistical anomalies such as Simpson's paradox~\cite{simpson1951interpretation} and Berkson's paradox~\cite{berkson1946limitations,kim1983computational}. A famous example of the Simpson's paradox is the gender bias in 1973 Berkley admission~\cite{berkeley75,loftus18}. In that year, 44\% of male applicants were admitted against only 34\% of female applicants. While this looks like a bias against female candidates, when the same data has been analyzed by department, acceptance rates were approximately the same. 

One way to address this limitation is to consider how data is generated in the first place which leads to causal-based fairness notions. Because this new breed of fairness notions is immune to statistical paradoxes, it is now widely accepted that causality is necessary to appropriately address the problem of fairness~\cite{loftus18}. Examples of causal-based fairness notions include total effect~\cite{pearl2009causality}, interventional fairness~\cite{salimi2019interventional}, counterfactual fairness~\cite{kusner2017counterfactual}, counterfactual effects~\cite{zhang2018fairness}, and path-specific counterfactual fairness~\cite{chiappa2019path,wu2019pc}. These notions are defined in terms of non-observable quantities such as causal, counterfactual, and path-specific effects. As they are non-observable, these quantities cannot always be estimated based on observable data. This is known as the \textit{identifiability} problem and is the topic of a large body of work in the causal inference literature. For example, the identifiability of causal effects can be decided using a set of three causal inference rules called do-calculus~\cite{pearl1995causal,pearl2009causality}. Based on the do-calculus, Shpitser and Pearl~\cite{shpitser2006identification} proposed a complete identification algorithm for causal effects. The algorithm (ID) was independently shown to be complete by Shpitser and Pearl~\cite{shpitser2006identification} and Huang and Valtorta~\cite{huang2006pearl}. Using the do-calculus for identifiability has two main issues. First, it is typically a manual process. Second, it is not clear in which order the rules should be applied~\cite{tikka2018thesis}. On the other hand, using the ID algorithm\footnote{Implemented in the \textbf{causaleffect} R package~\cite{tikka2017article1}.}, can produce unnecessarily complex expressions that can lead to inefficient, and even biased, estimates when data is missing feature values~\cite{tikka2017simplifying}. A more intuitive alternative for deciding about identifiability is to rely on graphical criteria, that is, recognizing common graph structures that produce identifiable effects. Graphical criteria is an efficient and intuitive approach to the identifiability of all types of effects (causal, counterfactual, and path-specific) and is more easy to use than the do-calculus or the identifiability algorithms (e.g. ID, ID*, etc.). 

This paper summarizes the main identifiability results as they relate to the specific problem of discrimination discovery with an emphasis on graphical criteria. These results fall into the following categories:
 causal effect (intervention) identifiability~\cite{galles1995testing,tian02,tian03,tian04,shpitser2006identification,huang06,shpitser08,pearl2009causality}, counterfactual identifiability~\cite{shpitser07-counterfactuals,shpitser08,shpitser13,wu19}, direct/indirect identifiability~\cite{pearl01direct}, and path-specific effect identifiability~\cite{avin2005identifiability,shpitser13,zhang2017anti,malinsky2019}. Section~\ref{preliminaries} provides necessary background concepts. Then, instead of repeating the definition of identifiability (Definition 3.2.3 in~\cite{pearl2009causality}), Section~\ref{sec:example} gives an intuitive explanation of the identifiability problem through the teacher firing example. Sections~\ref{sec:identcausaleffects}, \ref{sec:identctf}, and~\ref{sec:patheffect} compile the common identifiability results of causal, counterfactual, and path-specific effects, respectively.

\section{Preliminaries and Notation}
\label{preliminaries}
Variables are denoted by capital letters. In particular, $A$ is used for the sensitive variable (e.g., gender, race, age) and $Y$ is used for the outcome of the automated decision system (e.g., hiring, admission, releasing on parole). Small letters denote specific values of variables (e.g., $A=a'$, $W=w$). Bold capital and small letters denote a set of variables and a set of values, respectively.

A structural causal model~\cite{pearl2009causality} is a tuple $M = \langle \mathbf{U},\mathbf{V},\mathbf{F},P(\mathbf{U})\rangle$ where:
\begin{itemize}
	\item $\mathbf{U}$ is a set of exogenous variables which cannot be observed or experimented on but constitute the background knowledge behind the model.
	\item $\mathbf{V}$ is a set of observable variables which can be experimented on.
	\item $\mathbf{F}$ is a set of structural functions where each $f_i$ is mapping $\mathbf{U} \cup \mathbf{V} \rightarrow \mathbf{V}\backslash\{V_i\}$ which represents the process by which variable $V_i$ changes in response to other variables in $\mathbf{U} \cup \mathbf{V}$.
	\item $P(\mathbf{u})$ is a probability distribution over the unobservable variables $\mathbf{U}$. 

\end{itemize}

Causal assumptions between variables are captured by a causal diagram $G$ which is a directed acyclic graph (DAG) where nodes represent variables and directed edges represent functional relationships between the variables. Directed edges can have two interpretations. A probabilistic interpretation where the edge represents a dependency among the variables such that the direction of the edge is irrelevant. A causal interpretation where the edge represents a causal influence between the corresponding variables such that the direction of the edge matters. Unobserved variables $\mathbf{U}$, which are typically not represented in the causal diagram, can be either mutually independent (Markovian model) or dependent from each others. In case the unobserved variables can be dependent and each $U_i \in \mathbf{U}$ is used in at most two functions in $F$, the model is called semi-Markovian. In causal diagrams of semi-Markovian models, dependent unobservable variables (unobserved confounders) are represented by a dotted bi-directed edge between observable variables. Graphs $G5$ (Table~\ref{tab:markovian}) and $G16$ (Table~\ref{tab:semimarkovian}) show causal graphs of Markovian and semi-Markovian models, respectively.

\begin{figure}[!h]
\vspace{-2mm}
    \subfigure [Causal graph of a  Markovian model.]  {%
    {\includegraphics [scale=0.2]{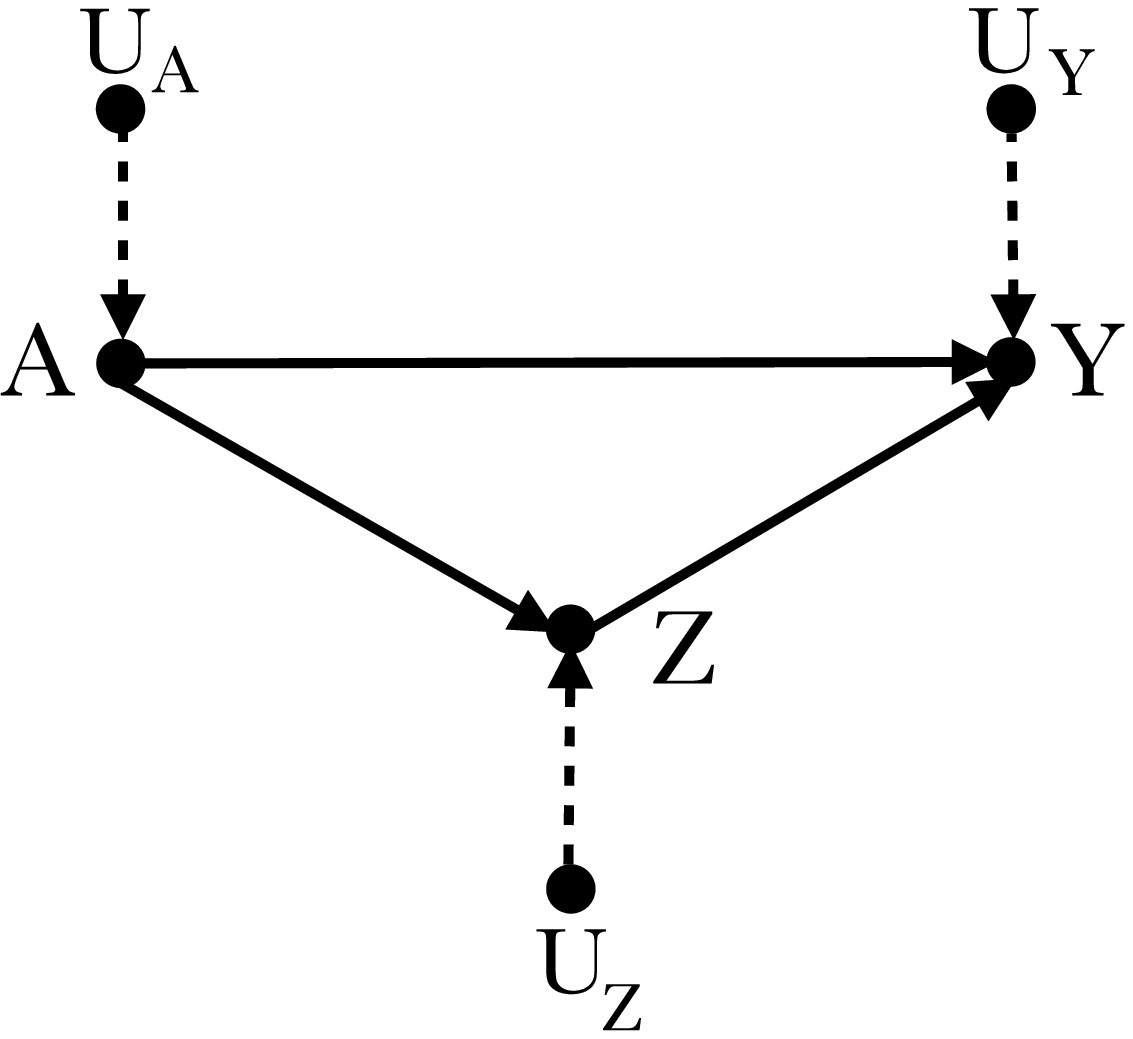} }
    \label{subfig:figMark_a}}
    \quad 
    \subfigure [Causal graph of a  semi-Markovian model.] {%
    {\includegraphics[scale=0.2]{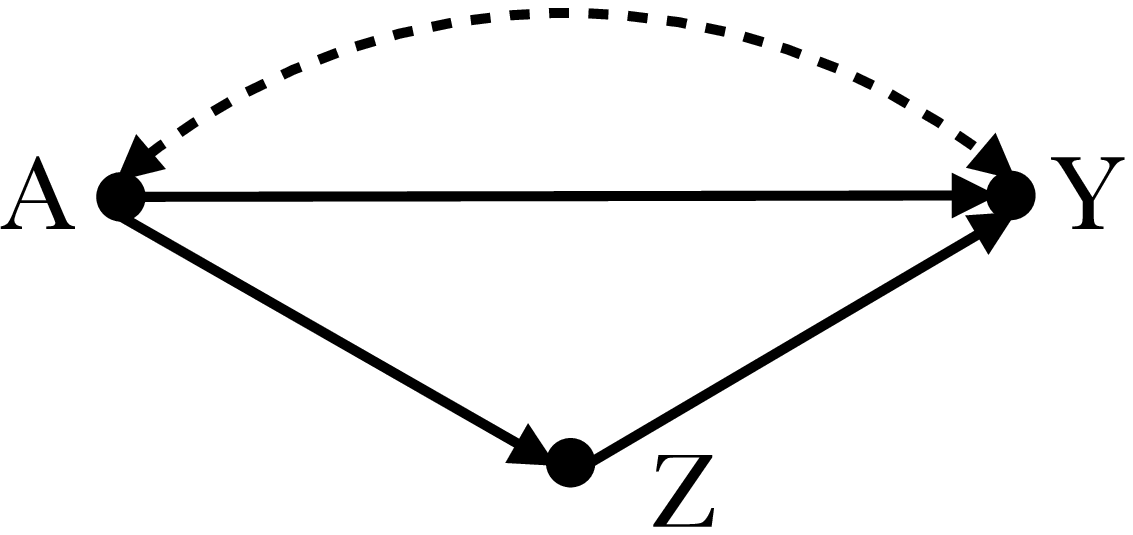} }
    \label{subfig:figMark_b}}
     \quad 
    \subfigure [Causal graph of a  semi-Markovian model after intervention: $do (Z = z)$.]  {%
    {\includegraphics [scale=0.2]{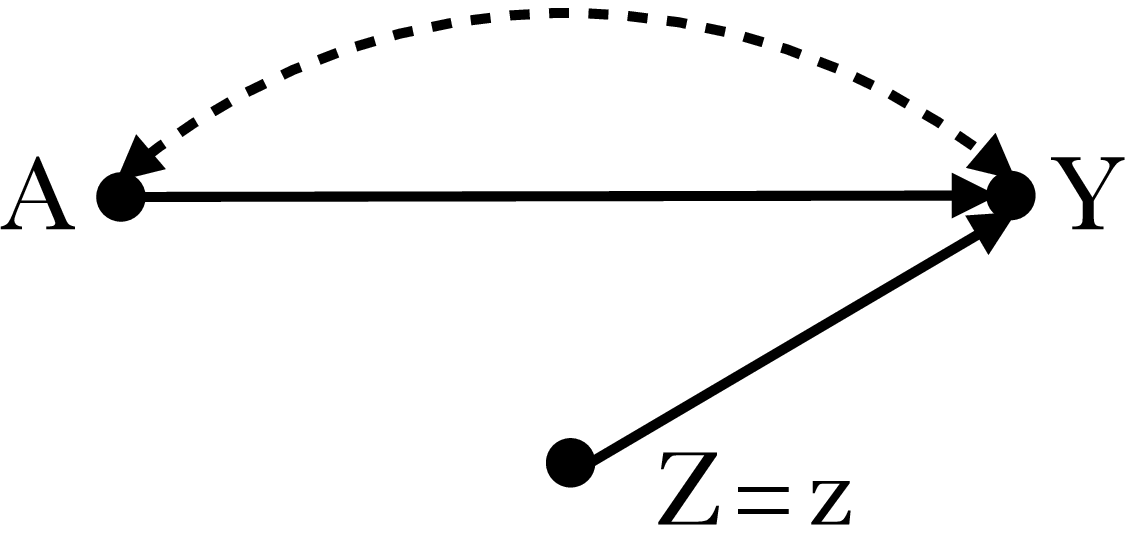} }
    \label{subfig:figMark_c}}
    \vspace{-2mm}
    \caption{}
    \label{fig:Mark_semiMark}
\end{figure}

An intervention, noted $do(V=v)$, is a manipulation of the model that consists in fixing the value of a variable (or a set of variables) to a specific value regardless of the corresponding function $f_v$. Graphically, it consists in discarding all edges incident to the node corresponding to variable $V$. Figure~\ref{subfig:figMark_c} shows the causal diagram of the manipulated model after intervention $do(Z=z)$ denoted $M_{Z=z}$ or $M_z$ for short. The intervention $do(V=v)$ induces a different distribution on the other variables. For example, in Figure~\ref{subfig:figMark_c}, $do(Z=z)$ results in a different distribution on $Y$, namely, $P(Y|do(Z=z))$. Intuitively, while $P(Y|Z=z)$ reflects the population distribution of $Y$ among individuals whose $Z$ value is $z$, $P(Y|do(Z=z)$ reflects the population distribution of $Y$ if \textit{everyone in the population} had their $Z$ value fixed at $z$. The obtained distribution $P(Y|do(Z=z)$ can be considered as a \textit{counterfactual} distribution since the intervention forces $Z$ to take a value different from the one it would take in the actual world. Such counterfactual variable is noted $Y_{Z=z}$ or $Y_z$ for short\footnote{The notations $Y_{Z\leftarrow z}$ and $Y(z)$ are used in the literature as well.}. $ P(Y=y | do(Z=z)) = P(Y_{Z=z} = y) = P(Y_z = y) = P(y_z)$ is used to define the causal effect of $z$ on $Y$. The term counterfactual quantity is used for expressions that involve explicitly multiple worlds. In Figure~\ref{subfig:figMark_b}, consider the expression $P(y_{a'}|Y=y, A=a) = P(y_{a'}|y,a)$. Such expression involves two worlds: an observed world where $A=a$ and $Y=y$ and a counterfactual world where $Y=y$ and $A=a'$ and it reads ``the probability of $Y=y$ had $A$ been $a'$ given that we observed $Y=y$ and $A=a$''. In the common example of job hiring, if $A$ denotes race ($a:$white, $a'$:non-white) and $Y$ denotes the hiring decision ($y$:hired, $y'$:not hired), $P(y_{a'}|y,a)$ reads ``given that a white applicant has been hired, what is the probability that the same applicant is still being hired had he been non-white''. 
Nesting counterfactuals can produce complex expressions. For example, in the relatively simple model of Figure~\ref{subfig:figMark_b}, $P(y_{a,z_{a'}}|y'_{a'}) = P( y(a,z(a'))|y'(a') )$ reads the probability of $Y=y$ had (1) $A$ been $a'$ and (2) $Z$ been $z$ when $A$ is $a'$, given that an intervention $A=a'$ produced $y'$. This expression involves three worlds: a world where $A=a$, a world where $Z=z_{a'}$, and a world where $A=a'$. Such complex expressions are used to characterize direct, indirect, and path-specific effects.

\section{Explaining identifiability through an example}
\label{sec:example}
Consider the example of an automated system for deciding whether to fire a teacher at the end of the academic year. Deployed teacher evaluation systems have been suspected of bias in the past. For example, IMPACT is a teacher evaluation system used in the city of Washington, D.C., and have been found to be unfair against teachers from minority groups~\cite{impact,impactBias,weapons16}. Assume that the system takes as input one feature, namely, the initial\footnote{At the beginning of the academic year.} average level of the students assigned to that teacher ($A$). The outcome is whether to fire the teacher ($\hat{Y}$). Assume that these two variables are confounded by a third unobservable variable $U$ which represents a socioeconomic status related to the school neighborhood.

Assume also that all 3 variables are binary with the following values: If the initial average level of the students assigned to the teacher is high, $A=1$, otherwise (initial level is low), $A=0$. Firing a teacher corresponds to $\hat{Y}=1$, while retaining her corresponds to $\hat{Y}=0$. If the school is located in a high-income neighborhood, $U=1$, otherwise (the school is located in a low-income neighborhood), $U=0$. The level of students in a given class can be influenced by several variables, but in this example, assume that it is only influenced by the socioeconomic status of the school; students in high-income neighborhoods are more advantaged and typically perform better in school. 

\begin{figure}[!h]
   \includegraphics [scale=0.3]{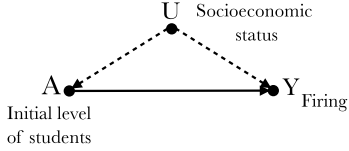} 
    \caption{Causal graph of the teacher firing example.}
    \label{fig:firingExample}
\end{figure}

The relationships between the variables $A,U,$ and $Y$ can be graphically represented using the causal directed acyclic graph (DAG) in Figure~\ref{fig:firingExample}\footnote{The structure of this graph is known as the bow structure in the literature.}. Notice that the edges $U \rightarrow A$ and $U\rightarrow Y$ are dotted because they are emanating from an unobservable variable ($U$). 

Assume that the automated decision system is suspected to be biased by the level of students assigned to the teacher. That is, it is claimed that the system is more likely to fire teachers who have been assigned classes with low level students at the beginning of the academic year, which is clearly unfair. The bias in the outcome ($\hat{Y}$) due to the sensitive variable $A$ can be assessed by computing the total variation:
\begin{equation}
\label{eq:TV}
TV_{a_1,a_0} (y) = P (y \mid a_1) - P (y \mid a_0)
\end{equation}
which coincides with statistical parity~\cite{dwork2012fairness} and     measures the difference between the conditional distributions of $\hat{Y}$ when we (passively) observe $A$ changing from $a_0$ to $a_1$ (e.g. from $0$ to $1$ in our example). The main limitation of $TV$ is that it is purely statistical and may be fooled by statistical anomalies such as Simpson's and Berkson's paradoxes. 
Total effect ($TE$)~\cite{pearl2009causality} is the causal version of $TV$ and is defined in terms of experimental probabilities as follows: 
\begin{align}
TE_{a_1,a_0} (y) & = P(Y=y | do(A=a_1)) - P(Y=y | do(A=a_0)) \nonumber \\
                 & = P (y_{a_1}) - P(y_{a_0}) \label{eq:TE}
\end{align}

While $TV$ is expressed in terms of observable probabilites ($P(y | a_1)$ and $P(y | a_0)$) and hence can always be computed from observable data, $TE$ is not. The question is can $TE$ be expressed in terms of observable probabilities and hence computed from observable data? If the answer is yes, $TE$ is said to be identifiable. Otherwise, it is not identifiable. Pearl gives a formal definition of identifiability ~\cite{pearl2009causality}, Page 77, Definition 3.2.3. Intuitively, given a dataset $D$ (which can be generated by different causal models), a quantity (e.g. $P(Y_{A=a_1} = y)$) is identifiable if it keeps the same value regardless of the causal model which generated the dataset $D$. For example, in the teacher firing scenario, $P(\hat{Y}_{A=0}=1)$ is not identifiable since it is possible to come up with two causal models that can generate the same data, and hence $P(\hat{Y}_{A=0}=1)$ cannot be uniquely computed based only on observable data. For illustration, consider the two following causal models ($\mathbf{M_1}$ and $\mathbf{M_2}$) expressed in terms of all three variables $A,\hat{Y},$ and $U$\footnote{$U$ is an exogenous variable and it is not observable by definition. But to illustrate the identifiability concept, assume there is an oracle with a knowledge about all model parameters including $U$.}:

\[\begin{array}{lcl}
        \text{\textbf{Causal model} }\mathbf{M_1}  & &    \\
		P(\hat{Y}=1\;|\;A=0, U=0) = 0.25 & & P(A=0\;|\;U=0) = 0.6\\
		P(\hat{Y}=1\;|\;A=1, U=1) = 0.25  & & P(A=0\;|\;U=1) = 0.4\\
		P(\hat{Y}=1\;|\;A=0, U=1) = 0.02 & & P(A=1\;|\;U=0) = 0.4\\
		P(\hat{Y}=1\;|\;A=1, U=0) = 0.02  & & P(A=1\;|\;U=1) = 0.6\\
\end{array}\]

\[\begin{array}{lcl}
        \text{\textbf{Causal model} }\mathbf{M_2}  & &    \\
		P(\hat{Y}=1\;|\;A=0, U=0) = 0.24 & & P(A=0\;|\;U=0) = 0.65\\
		P(\hat{Y}=1\;|\;A=1, U=1) = 0.24  & & P(A=0\;|\;U=1) = 0.35\\
		P(\hat{Y}=1\;|\;A=0, U=1) = 0.01 & & P(A=1\;|\;U=0) = 0.35\\
		P(\hat{Y}=1\;|\;A=1, U=0) = 0.01 & & P(A=1\;|\;U=1) = 0.65\\
\end{array}\]

It is easy to show that both causal models generate the same joint distribution $P(\hat{Y},A)$. Using the chain rule, 
\begin{align}
	P(A,\hat{Y}) & = \sum_{u\in\{0,1\}} P(\hat{Y}|\;A,U=u) P (A |\; U=u) P(U=u) \nonumber \\
\end{align}
both $\mathbf{M_1}$ and $\mathbf{M_2}$ generate the same observable distribution:
\[\begin{array}{l}
		P(\hat{Y}=1\;,\;A=1) = 0.079  \\
		P(\hat{Y}=0\;,\;A=0) = 0.42  \\
		P(\hat{Y}=1\;,\;A=0) = 0.079\\
		P(\hat{Y}=0\;,\;A=1) = 0.42 \\
\end{array}\]

$P(\hat{Y}_{A=0}=1)$ is not an observable quantity. However, since we assumed the existence of an oracle with knowledge about all model parameters, it can be computed using the back-door formula (Equation~\ref{eq:back}) as follows:
\begin{align}
	P(\hat{Y}_{A=0}=1) & = \sum_{u\in\{0,1\}} P(\hat{Y}=1|\;A=0,U=u) P(U=u) \nonumber \\
\end{align}

For causal model $\mathbf{M_1}$, 
\begin{align}
    P(\hat{Y}_{A=0}=1) = (0.25 \times 0.5) + (0.02 \times 0.5) = 0.135 \nonumber 
\end{align}
whereas for causal model $\mathbf{M_2}$, 
\begin{align}
    P(\hat{Y}_{A=0}=1) = (0.24 \times 0.5) + (0.01 \times 0.5) = 0.125  \nonumber 
\end{align}

Hence, $\mathbf{M_1}$ and $\mathbf{M_2}$ are two different causal models that generate the same observable data but yield two different values for the quantity $P(\hat{Y}_{A=0}=1)$ which is consequently not identifiable from observational data. In other words, in this situation, it is not possible to use observations to tell whether $A$ is actually a cause of $\hat{Y}$.

Since total variation $TV$ is defined in terms of observable probabilities, it can be computed based on the observable data. Total effect $TE$, however, cannot be computed based on observable data as $P(\hat{Y}_{A=0}=1)$ is not identifiable. 

Notice that, in this example, both models $\mathbf{M_1}$ and $\mathbf{M_2}$ share the same graph structure (Figure~\ref{fig:firingExample}). This is not always the case. That is, it is possible to have two causal models with different graph structures coinciding on the observable joint distribution. Hardt et al.~\cite{hardt2016equality} illustrate this case with an example. Tikka~\cite{tikka2018thesis} presents another non-identifiable example defined using the XOR logic operator.

Based on the causal inference literature, the next sections compile a list of identifiability criteria for the different types of non-observable quantities: causal, counterfactual, direct, indirect, and path-specific effects.

\section{Identifiability of causal effects}
\label{sec:identcausaleffects}

 The natural way to estimate the causal effect of a variable (the sensitive attribute $A$) on another (the outcome variable $Y$) is to carry out real experiments using RCT (Randomized Controlled Trial)~\cite{fisher92}. If possible, RCT drops the need for identiability altogether. However, in the context of machine learning fairness, RCT is often not an option as experiments can be too costly to implement or physically impossible to carry out (e.g. changing the gender of a job applicant). 

As an alternative, intervention using the do-operator can be used to compute the causal effect. Without loss of generality, this section focuses on the identifiability of $P(Y=y | do(A=a)) = P(y_a)$, that is, the causal effect of the sensitive attribute $A$ on the outcome variable $Y$. The computation of $P(y_a)$ uses a ``surgically altered'' graph in which all arrows into $A$ are deleted and the value of $A$ is fixed at $a$, but the rest of the graph remains unchanged. 

Whether it is possible to express $P(y_a)$ only in terms of observable probabilities (identifiability)  depends on the structure of the causal graph (which captures how data is generated).  A first important result is that any causal effect is identifiable in a Markovian model (where all unobservable variables are independent). In semi-Markovian models, however, the causal effect is not always identifiable. 


\subsection{Identifiability in Markovian models}
\label{subsec:Markov}


Table~\ref{tab:markovian} shows different Markovian models involving various patterns of causal relationships along with the corresponding expression in terms of observable probabilities.

\begin{table}[!h]
\setlength\extrarowheight{3pt} 
    \begin{tabular}{|c|c|l|}
    \hline
         & Causal graph & \makecell{$P(y_a)$} \\
    \hline
       \makecell{$G1$\\\\}  & \includegraphics [scale=0.2]{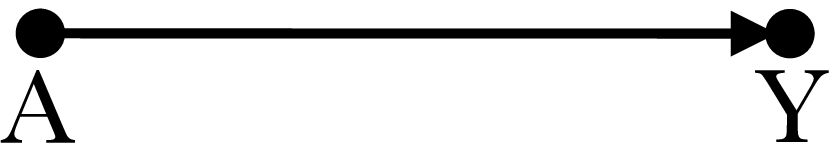}\label{G1}  & \\
        \cline{1-2}
       \makecell{$G2$\\\\}  & \includegraphics [scale=0.2]{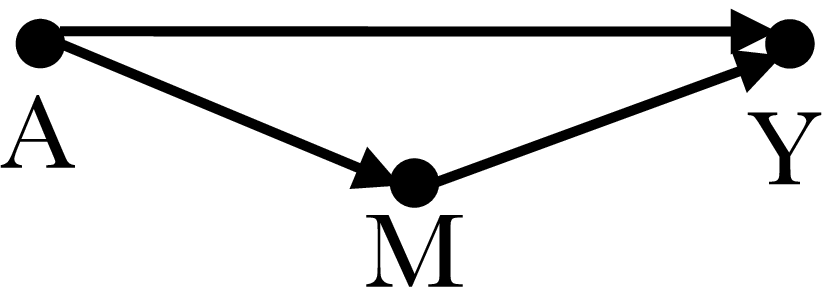} &\makecell{{\multirow{5}{*}{$P(y|a)$}}} \\
        \cline{1-2}
       \makecell{$G3$\\\\}  & \includegraphics [scale=0.2]{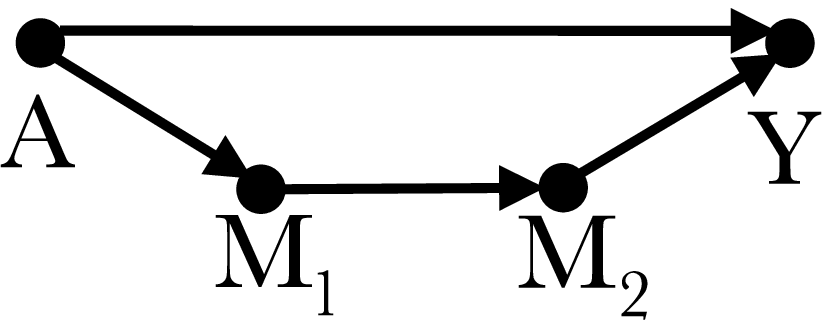} & \\
        \cline{1-2}
       \makecell{$G4$\\\\}  & \includegraphics [scale=0.2]{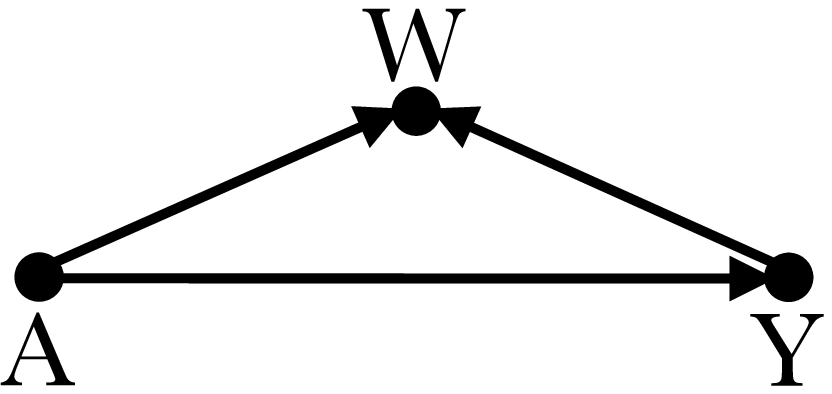} & \\
        \cline{1-2}
       \makecell{$G5$\\\\\\}  & \includegraphics [scale=0.2]{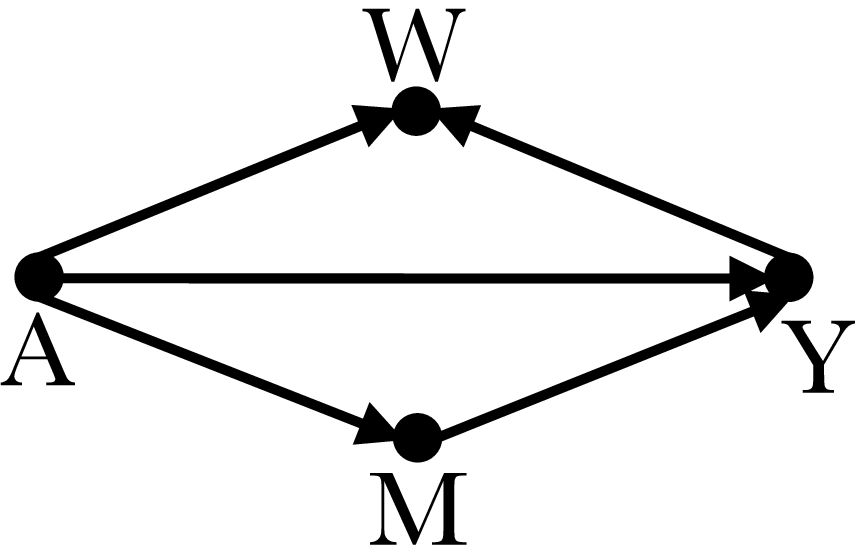} & \\
        \hline
       \makecell{$G6$\\\\}  & \includegraphics [scale=0.2]{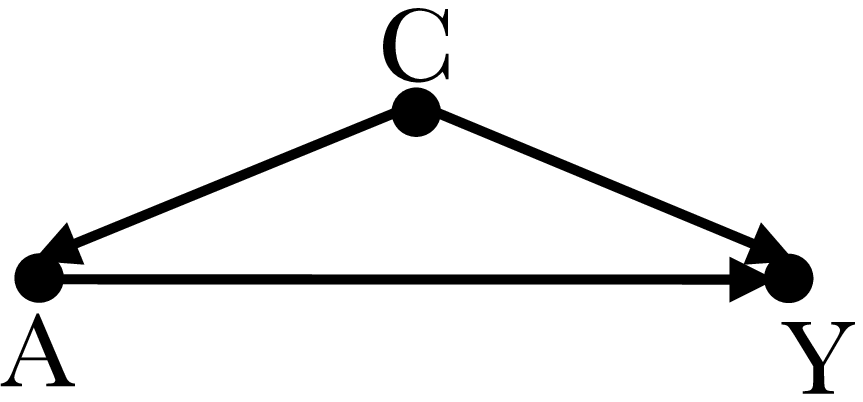} & \makecell{{\multirow{4}{*}{$\sum_{C}P(y|a,c)\;P(c)$}}} \\
        \cline{1-2}
        \makecell{$G7$\\\\\\}  & \includegraphics [scale=0.2]{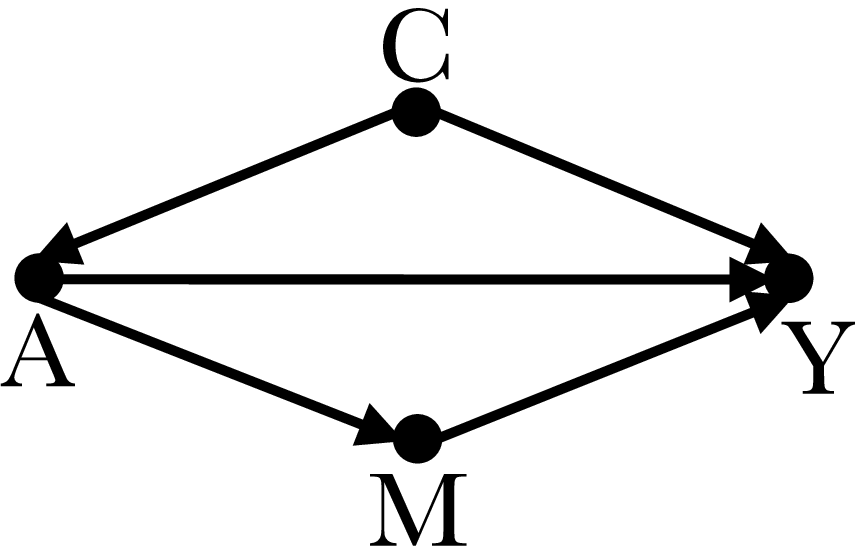}& \\
        \cline{1-2}
       \makecell{$G8$\\\\\\}  & \includegraphics [scale=0.2]{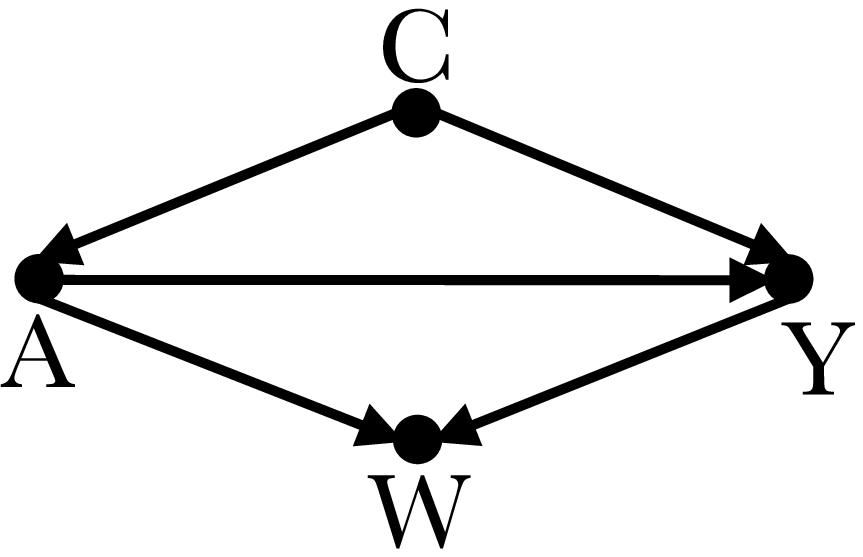} & \\
           \cline{1-2}
       \makecell{$G9$\\\\\\}  & \includegraphics [scale=0.2]{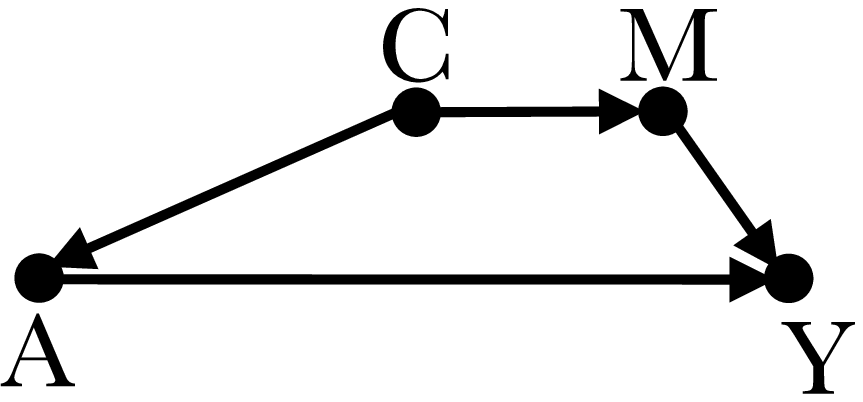} & \\
       \hline
       \makecell{$G10$\\\\\\}  & \includegraphics [scale=0.2]{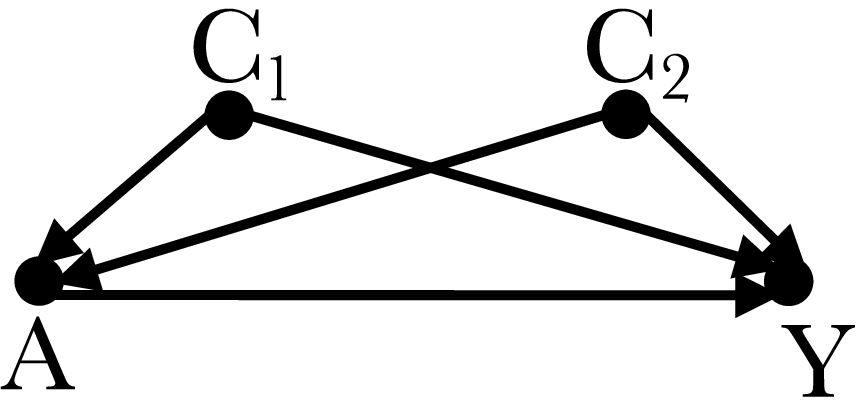} & $\sum_{C_1 C_2}P(y|a,c_1,c_2)\;P(c_1,c_2)$\\
        \hline
       {\multirow{6}{*}{$G11$}}  & {\multirow{6}{*}{\includegraphics [scale=0.2]{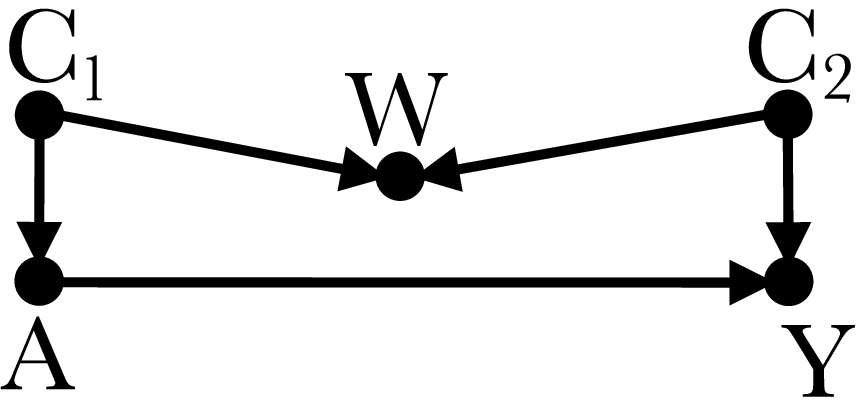}}} & $P(y|a)$\\
        & & $\sum_{C_1}P(y|a,c_1)\;P(c_1)$\\
        &  & $\sum_{C_2}P(y|a,c_2)\;P(c_2)$\\
         &  & $\sum_{W C_1}P(y|a,w,c_1)\;P(w,c_1)$\\
          &  & $\sum_{W C_2}P(y|a,w,c_2)\;P(w,c_2)$\\
          &  & $\sum_{W C_1 C_2}P(y|a,w,c_1,c_2)\;P(w,c_1,c_2)$\\
        \hline
       {\multirow{4}{*}{$G12$}}  & {\multirow{4}{*}{\includegraphics [scale=0.2]{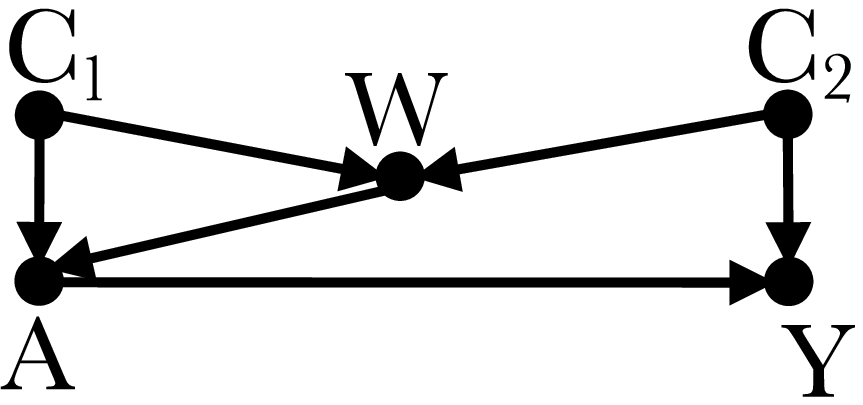}}} & $\sum_{C_2}P(y|a,c_2)\;P(c_2)$\\
        & & $\sum_{W C_1}P(y|a,w,c_1)\;P(w,c_1)$\\
        &  & $\sum_{W C_2}P(y|a,w,c_2)\;P(w,c_2)$\\
         &  & $\sum_{W C_1 C_2}P(y|a,w,c_1,c_2)\;P(w,c_1,c_2)$\\
        \hline
    \end{tabular}
    \caption{$P(y_a)$ of some Markovian models.}
    \label{tab:markovian}
\end{table}

Graphs $G1-G5$ illustrate the simplest cases where no confounding between $A$ and $Y$ exists. In that case, the causal effect matches the conditional probability regardless of any mediator $M$ as follows:
\begin{equation}
\label{eq:simple}
	P(y_a) = P(y|a) 
\end{equation}

\subsubsection{Back-door adjustment}
\label{subsec:back}

In case there are confounders involving $A$ and $Y$, the causal effect can be identified by finding a set of variables $C$ that block all back-door paths from $A$ to $Y$. This is called the back-door criterion\footnote{Called also adjustment formula or stratification.}. This criterion necessitates the existence of a set of covariates $C$ which blocks all the indirect paths from $A$ to $Y$, but keeps all the direct paths open. $C$ satisfies the back-door criterion when (1) $C$ blocks every back-door path between $A$ and $Y$, and (2) no node in $C$ is a descendant of $A$. Graphs $G6-12$ illustrate examples where $C$ (or \{$C_1$,$C_2$\}) meets the back-door criterion. In presence of an observable confounder $C$, $P(y_a)$ is identifiable by adjusting\footnote{The terms adjusting, controlling, and marginalizing are used interchangeably.} on that confounder using back-door formula:
    \begin{equation}
        \label{eq:back}
        P(y_a) = \sum_{C}P(y|a,c)\;P(c) 
    \end{equation}
     where the summation is on values $c$ in the domain (sample space) of $C$ denoted as $dom(C)$. 
Note that $G4$ and $G5$ contain a collider ($W$). Marginalizing over the collider variable disproves the equality in Eq.~\ref{eq:back} as it might open back-door paths between $A$ and $Y$ and consquently create a dependency between these two variables. 

Despite the fact that $G11$ involves two confounders $C_1$ and $C_2$, no adjustment is required because of the presence of the collider $W$. Hence $P(y_a)$ can be computed using Eq.~\ref{eq:simple}. Alternatively, controlling on: $C_1$, $C_2$, $\{W,C_1\}$,$\{W,C_2\}$ or $\{W,C_1,C_2\}$ is possible using Eq.~\ref{eq:back}. Table~\ref{tab:markovian} shows all possible formulas that can be used to calculate $P(y_a)$ for $G11$. $G12$ presents another case with two confounders ($C1$ and $C_2$) and the two following back-door paths between $A$ and $Y$: $A \leftarrow W \leftarrow C_2  \rightarrow Y$ and $A \leftarrow C_1 \rightarrow W  \leftarrow C_2 \rightarrow Y$. The former must be blocked by either $W$ or $C_2$ or both while the latter doesn't need any controlling because of the presence of the collider: $W$. Thus, the set of variables sufficient to control for confounding are: $C_2$, $\{C_1,W\}$, $\{W,C_2\}$ or $\{W,C_1,C_2\}$ but not $W$ or $C_1$ (the minimum to control for is: $C_2$). That is, any one of these equations can be used to calculate the causal effect of $A$ on $Y$.

\subsubsection{Truncated factorization formula}
\label{subsec:trunc}
An alternative way to measure the causal effect $P(y_a)$ in Makovian models is to use the truncated factorization formula~\cite{pearl2009causality}:
    \begin{equation}
    \label{eq:trunc}
	P(y_a) = \sum_{\substack{\mathbf{V}\backslash\{A,Y\}\\ Y=y}} \prod_{V \in \mathbf{V}\backslash\{A\}} P(v|{\mathbf{Pa}}_{V}) 
    \end{equation}
    where $\mathbf{Pa}_{V}$ denotes the parent variables of $V$.
For instance, applying the truncated factorization formula on $G12$ leads to the following equality:
\begin{align}
    P(y_a) = \sum_{W C_1 C_2}P(y|\;a,c_2)\;P(w|\;c_1,c_2)\;P(c_1)\;P(c_2)
\end{align}
Note that Eq.~\ref{eq:back} and the last result of applying the back-door criterion for $G12$ in Table~\ref{tab:markovian} are equivalent. This can be easily demonstrated as follows:
\begin{align}
    & \sum_{W C_1 C_2} P(y|a,w,c_1,c_2)\;P(w,c_1,c_2) \nonumber \\
    = &\sum_{W C_1 C_2} P(y|a,w,c_1,c_2)\;P(w|\;c_1,c_2)P(c_1)\;P(c_2) \\
    = &\sum_{W C_1 C_2} P(y|a,c_2)\;P(w|\;c_1,c_2)P(c_1)\;P(c_2) 
\end{align}
$P(y|a,w,c_1,c_2)$ in (9) is replaced by $P(y|a,c_2)$ in (10) due to the fact that $Y\!\perp\!\!\!\perp W |\;A$\footnote{$Y$ and $W$ are independent given $A$.} and $Y\!\perp\!\!\!\perp C_1 |\;A$.

As a summary, the only type of variables that have an impact on the identifiability of $P(y_a)$ in Markovian models is the confounder. To compute the causal effect in presence of confounding, adjusting using the back-door formula (Eq.~\ref{eq:back}) is required.  However, adjusting should not be used in presence of a collider variable since this might open back-door paths between $A$ and $Y$ and hence, create a dependency between them. Mediator variables, on the other hand, have no impact on the identifiability of causal effects in Markovian models.

\subsection{Identifiable semi-Markovian models}
\label{subsec:identsemiMarkov}       

Causal effects are not always identifiable in semi-Markovian models. This subsection focuses on causal models where the causal effect of $A$ on $Y$ is identifiable. The following subsection gives a graphical criteria of causal models where the causal effect is not identifiable. 


In the causal model, the measurement of causal effects is assisted by interventions following a set of inference rules introduced by Pearl~\cite{pearl2009causality} known as: \textit{do-calculus}. These rules tend to link the interventional quantities of causal effects to simple statistical distributions based solely on observational data. As an alternative way of assessing causal effects, relevant graphical patterns will be presented in the remainder of this section. 

\subsubsection{do-calculus inference rules}
\label{subsec:calculus}
do-calculus~\cite{pearl1995causal,pearl2009causality} is a set of three inference rules that can be used to express an interventional expression of the form $P(y_a)$ in terms of subscript-free (observable) quantities. The rules are: 



\begin{itemize}
    \item \textbf{Rule 1 (Insertion/Deletion of Observations):}\\
    $P(y_a|\mathbf{c},w) = P(y_a|\mathbf{c})$ provided that the set of variables $\mathbf{C}$ blocks all back-door paths from $W$ to $Y$ after all arrows leading to $A$ have been deleted.
    \item \textbf{Rule 2 (Action/Observation Exchange):}\\
    $P(y_a|\mathbf{c}) = P(y|a,\mathbf{c})$ provided that the set of variables $\mathbf{C}$ blocks all back-door paths from $A$ to $Y$. 
    \item \textbf{Rule 3 (Insertion/Deletion of Actions):}\\
    $P(y_a) = P(y)$ provided that there are no causal paths between $A$ and $Y$.
\end{itemize}

Although the do-calculus is proven to be complete for identifying causal effects\footnote{If an interventional expression cannot be converted into subscript-free quantity, it means the expression is not identifiable.}~\cite{huang2006pearl,shpitser2006identification}, the completeness is not immediately apparent from the rules themselves. The other issue is that do-calculus is typically used manually and hence it is not obvious in which order the rules should be used to reach the subscript-free expression. An example of using the do-calculus is detailed in~\cite{pearl2009causality} Section 3.4.3. Deciding about the identifiability of causal effect is not easy with the do-calculus. A more intuitive approach would be to use graphical criteria. The rest of the subsection lists the most common graphical criteria for the identifiability of causal effects in semi-Markovian models.

\subsubsection{Graphical criteria}
\label{graphs}
The simplest case where the causal effect of $A$ on $Y$ is identifiable in semi-Markovian models is when $A$ is not connected to any unobserved confounder, that is, no bi-directed edge is connected to $A$. Graphs $G13$ and $G14$ in Table~\ref{tab:semimarkovian} satisfy this criterion. Now, depending on the existence (or absence) of back-door paths connecting $A$ to $Y$, the calculation of the causal effect varies. Then, in case all the pathways connecting $A$ to $Y$ are front-door from $A$ (start with an outgoing edge from $A$), the causal effect coincides with the conditional probability (Eq.~\ref{eq:simple}). $G13$ illustrates an example of such situation. On the other hand, in case $A$ is connected to some observed confounders (there is a pathway from $A$ to $Y$ staring with an edge into $A$), the back-door formula (Eq.~\ref{eq:back}) is needed to compute the causal effect. $G14$ presents such pattern. This matches Theorem~3.2.5 in~\cite{pearl2009causality} which states that if all parents of a cause variable $A$ are observable, the causal effect of that variable is identifiable. Hence, the back-door formula can be generalized as follows (Theorem~3.2.5~\cite{pearl2009causality}):
\begin{equation} 
    \label{eq:ident1}
     P(y_a) = \sum_{pa_A} P(y|a,\mathbf{pa}_A)\;P(\mathbf{pa}_A) 
\end{equation}
where $\mathbf{pa}_A$ is the set of values of the parents of $A$. $G16$ is a more complex causal model that satisfies Eq.~\ref{eq:ident1} where the causal effect of $A$ on $Y$ is identifiable and can be computed as:
\[
 P(y_a) = \sum_{c_1,c_2} P(y|a,c_1,c_2)\;P(c_1,c_2)
\]
 
 \begin{table}[!h]
\setlength\extrarowheight{15pt} 
    \begin{tabular}{|c|c|l|}
    \hline
         & Causal graph & \makecell{$P(y_a)$} \\
    \hline
       \makecell{$G13$\\\\}  & \includegraphics [scale=0.2]{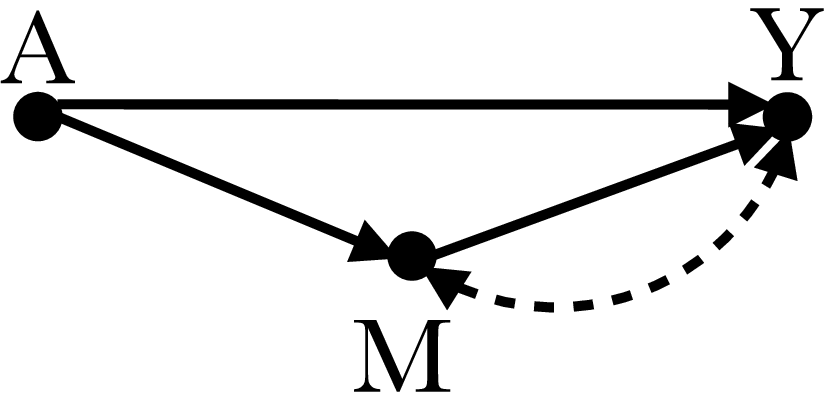} & \makecell{$P(y|a)$}\\
    \hline
       \makecell{$G14$\\\\}  & \includegraphics [scale=0.2]{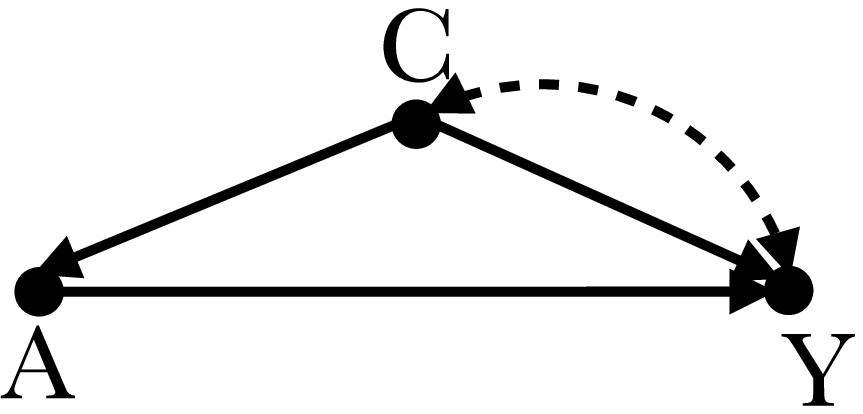} & \makecell[l]{{\multirow{2}{*}{$\sum_{C}P(y|a,c)\;P(c)$}}}\\
    \cline{1-2}
       \makecell{$G15$\\\\}  & \includegraphics [scale=0.2]{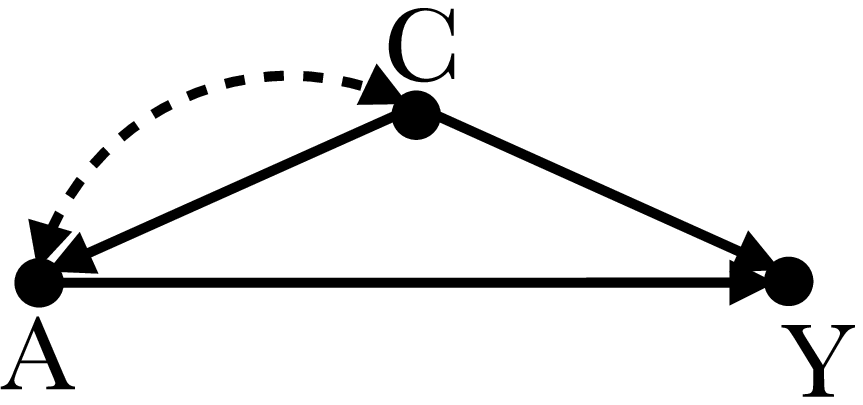} & \\
    \hline
       {\multirow{2}{*}{$G16$}}  & {\multirow{2}{*}{\includegraphics [scale=0.2]{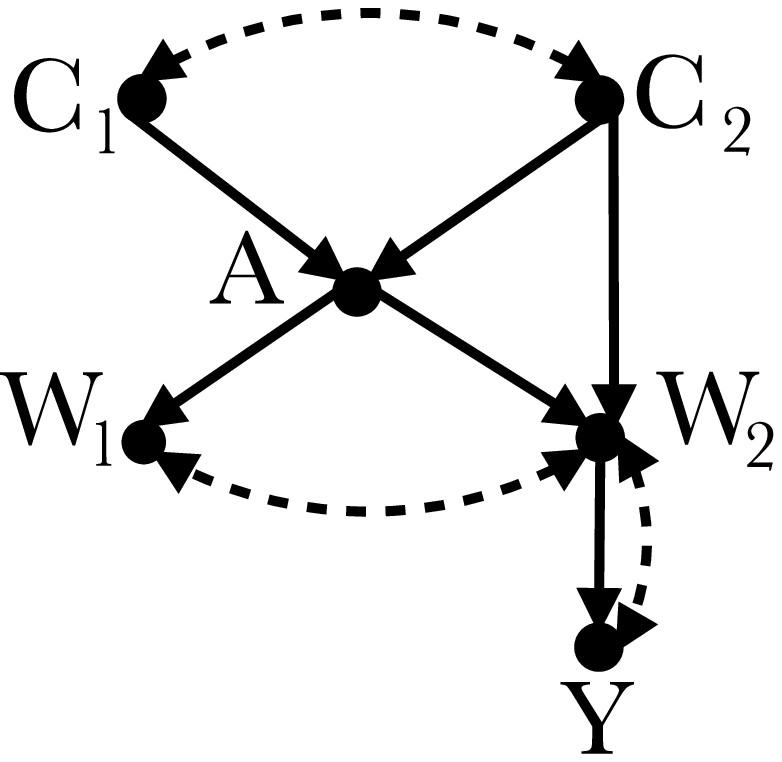}}} & \makecell[l]{$\sum_{c_1,c_2} P(y|a,c_1,c_2)\;P(c_1,c_2)$}\\
       &  & \\
    \hline
       {\multirow{2}{*}{$G17$}}  & {\multirow{2}{*}{\includegraphics [scale=0.2]{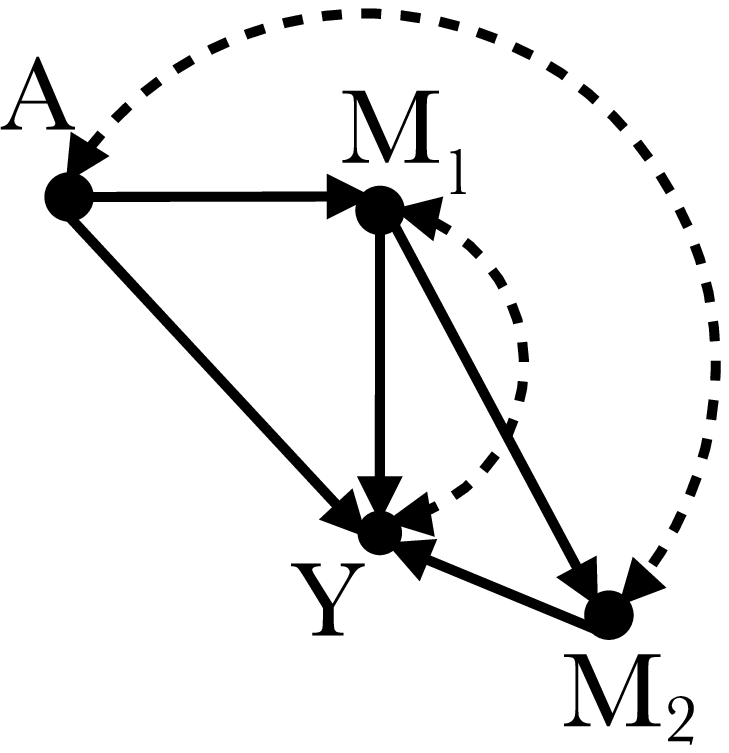}}} & $\sum_{m_1,m_2}\; P(y|m_1, m_2,a)\;P(m_1|a)$\\
    &  & \makecell[l]{$\times \sum_{a'}\;P(m_2|m_1,a')\;P(a')$\\\\}\\
    \hline
       {\multirow{2}{*}{$G18$}}  & {\multirow{2}{*}{\includegraphics [scale=0.2]{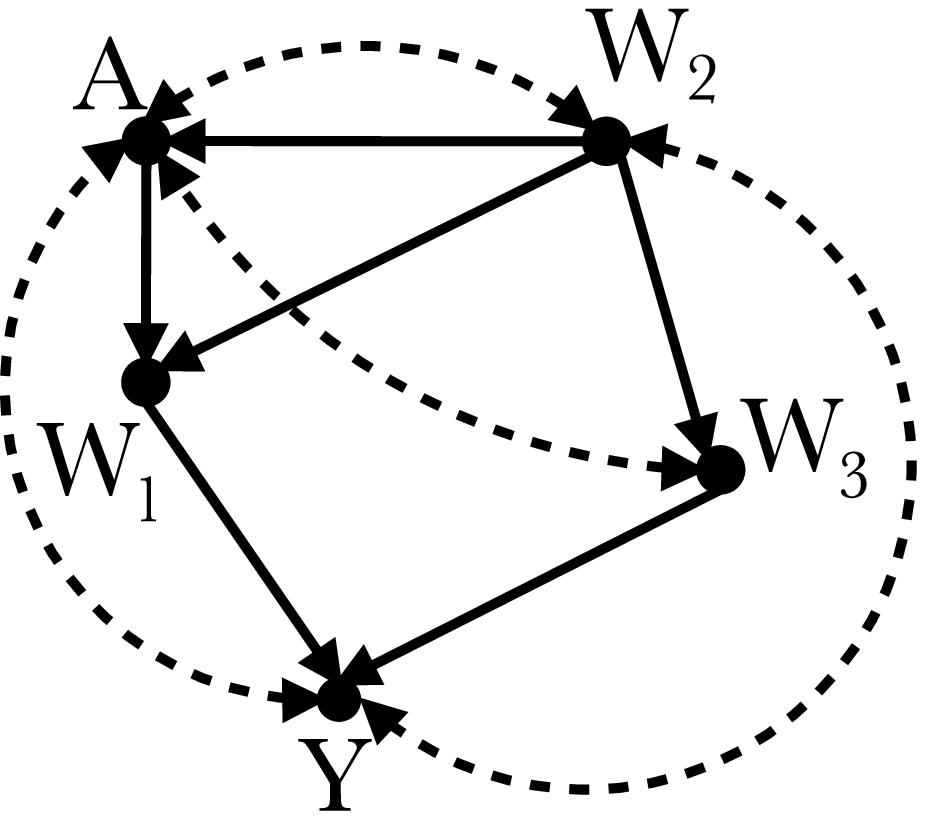}}} & $\sum_{w_1} \sum_{w_2} \sum_{a'}\; P(y|w_1,w_2,a')$ \; \\
    &  & \makecell[l]{$\times \;P(a'|w_2) \; P(w_1|w_2,a) P(w_2)$\\\\}\\
    \hline
       {\makecell{$G19$\\\\}}  & \makecell{{\includegraphics [scale=0.55]{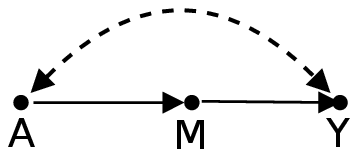}}} & \makecell[l]{$\sum_{m} \;P(m|a)\;\sum_{a'}P(y|m,a')\;P(a') $\\\\} \\
    \hline
    \end{tabular}
    \caption{$P(y_a)$ of some semi-Markovian models.}
    \label{tab:semimarkovian}
\end{table}
 Note that, in $G15$, despite the fact that $A$ is involved in confounding, Eq.~\ref{eq:back} also applies as $C$ blocks all the back-door paths including the unobserved one: $A \dashleftrightarrow C$. 
 
$G17$ and $G18$ are more complex causal models, but still identifiable due to a more specific graphical criterion: $A$ is not connected through bi-directed and dashed paths to any of its children that are at the same time ancestors of $Y$. Under such criterion, the causal effect of a single variable $A$ on all the other variables $\mathbf{V}\backslash\{A\}$ in the model denoted as $P(\mathbf{v}_a) = P_a(\mathbf{v})$ is identifiable and is given by Theorem 2~\cite{tian02}:
\begin{equation}
\label{eq:ident2}
	P_a(\mathbf{v}) = \left( \prod_{i|V_i \in \mathbf{ch_A}} P(v_i|\mathbf{pa}_i) \right) \sum_{a'\in dom(A)} \frac{P(v)}{ \prod_{i|V_i \in \mathbf{ch_A}} P(v_i|\mathbf{pa}_i) } 
\end{equation}
where $ch_A$ is the set of the children of the node $A$ while $\mathbf{pa_i}$ is the set of values of the parents of the variable $V_i$.

Now, since our aim is to assess the effect of the sensitive attribute $A$ on a single variable (the outcome $Y$), Eq.~\ref{eq:ident2} should be adapted. To illustrate that, consider the example of the causal graph $G18$. Applying Eq.~\ref{eq:ident2} to $G18$ leads to:

\begin{align}
    P_a(w_1,w_2,w_3,y) = & P(w_1|\;a,w_2)\sum_{a'\in dom(A)} \frac{P(w_1,w_2,w_3,y)}{ P(w_1|\;a',w_2)}) \label{eq:tian1}\\
    = & P(w_1|\;a,w_2) \sum_{a'\in dom(A)} P(y,w_3|\;a',w_1,w_2) \nonumber\\
    & \times P(a',w_2) \label{eq:tian2} 
\end{align}
where Eq.~\ref{eq:tian2} is obtained by applying the Bayes' rule.
Adapting Eq.~\ref{eq:ident2} in order to compute $P(y_a)$ (causal effect on the single variable $Y$) requires summing over the possible values of variables $W_1$, $W_2$ and $W_3$ as follows. Starting from Eq.~\ref{eq:tian2}, summing over $W_1$ gives:
\begin{align}
    P_a(w_2,w_3,y) = & \sum_{w_1'\in dom(W_1)} P(w_1'|\;a,w_2) \nonumber \\
    & \times \; \sum_{a'\in dom(A)} P(y,w_3|\;a',w_1',w_2) P(a',w_2) \nonumber
\end{align}

Similarly, summing over $W_2$ and $W_3$ (and omitting $dom()$ for conciseness), leads to:
\begin{align}
    P(y_a)  & = \sum_{w_1'}\; \sum_{w_2'}\; \sum_{a'}\; P(y|w_1',w_2',a')\;P(a'|w_2') \; P(w_1'|w_2',a)\;P(w_2') \label{eq:tian3}
\end{align}

Note that $W_3$ is omitted from Eq.~\ref{eq:tian3} as this variable is not concerned by the causal effect of $A$ on $Y$~\cite{pearl2009causality}.

Pearl~\cite{pearl2009causality} obtained the same result (Eq.~\ref{eq:tian3}) using \textit{do-calculus} (Section~\ref{subsec:calculus}). Thus, 
\begin{align}
    P(y_a)  & = \sum_{w_1}\; \sum_{w_2}\; \sum_{w_3}\; P(y|w_1,w_3)\;P(w_2)\;P(w_1,y|\;do(a)) \label{eq:pearl1} \\
     & = \sum_{w_1}\; \sum_{w_2}\; P(y|w_1,w_2)\;P(w_2)\;P(w_1,y|\;do(a)) \label{eq:pearl2} \\
     & = \sum_{w_1}\; \sum_{w_2}\;     P(y|w_1,w_2)\;P(w_2)\;P(w_1|\;a,w_2) \nonumber \\
     & \times \; \sum_{a'} P(y|\;w_1,a')\;P(a'|\;w_2) \label{eq:pearl3} \\
      & = \sum_{w_1}\; \sum_{w_2}\;\sum_{a'}  P(y|w_1,w_2,a')\;P(w_2)\;P(w_1|\;a,w_2)\;P(a'|\;w_2) \label{eq:pearl4}
\end{align}
Note that Eq.~\ref{eq:pearl4} is exactly the same as Eq.~\ref{eq:tian3}. For the same reason stated earlier, $w_3$ is ommitted in (17). The term $P(w_1,y|\;do(a))$ in (17) is replaced by $P(w_1|\;a,w_2) \sum_{a'} P(y|\;w_1,a')\;P(a'|\;w_2)$ after applying Rule 2 followed by Rule 3 of \textit{do-calculus} (symbolic derivation of Causal Effects: Eq. 3.43~\cite{pearl2009causality}). Since $W_2$ blocks all back-door paths between $A$ and $Y$, we apply the back-door formula (Eq.~\ref{eq:back}) to adjust on $W_2$ in (19).

\subsubsection{C-component factorization}
\label{subsec:c-comp}
C-component factorization~\cite{shpitser08} aims to express the observational distribution $P(v)$ as a product of factors $P_{v\setminus s}(s)$, where each $s$ represents the set of vertices included in a c-component. A c-component is a set of vertices in the graph such that every pair of vertices are connected by a confounding edge. The c-components are very important in measuring the causal effect of $A$ on $Y$ since they help in decomposing the identification problem into smaller sub-problems.
In other words, variables in the graph can be partitioned into a disjoint set of c-components in order to calculate $P(y_a)$. For example, the graph $G17$ is partitioned into two c-components: $S_1=\{A,M_2\}$ and $S_2=\{M_1,Y\}$ while the c-components of $G18$ are: $S_1=\{A,W_2,Y,W_3\}$ and $S_2=\{W_1\}$. Shpitser and Pearl~\cite{shpitser08} designed an algorithm called \textbf{ID} which aims to decompose the identification problem into smaller sub-problems based on the \textit{c-component factorization} property. This algorithm provides a complete solution for computing all identifiable causal effects.

Note that as long as there is no confounding path connecting $A$ to any of its direct children, $P(y_a)$ is identifiable and can be computed as~\cite{tian02}:
\begin{equation}
\label{eq:component}
	P(y_a) = \frac{P(y)}{Q^{A}} \sum_{a'}Q^{A}
\end{equation}
where $Q^{A}$ is the c-factor of the c-component containing A ($S^A$) computed as follows:
\begin{equation}
       Q^A = \prod_{v\in S^A} P(v|\mathbf{v}^{-1})
\end{equation}

where $\mathbf{v}^{-1}$ is the set of values of all previous variables to $V$, assuming a topological order $V_1\prec V_2 \prec \ldots \prec V_n$ over all variables. For instance, in $G17$, $W_2 \prec A \prec M_1 \prec M_2 \prec Y$ is a valid topological order. This criterion can be slightly generalized to be: $P(y_a)$ is identifiable if there is no confounding path connecting $A$ to any of its children in $G_{An(Y)}$ which is the subgraph of $G$ composed only of ancestors of the outcome variable $Y$. 

To illustrate the c-component factorization property, consider the causal graph $G17$.
Hence, applying Eq.~\ref{eq:component} to $G17$ leads to:
\begin{align}
    P(y_a) = \frac{P(y)}{P(a)P(m_2|\;a,m_1)} \sum_{a'}P(a')P(m_2|\;a',m_1)
\end{align}


Note that the causal effect $P(y_a)$ is not identifiable in a causal graph composed of a single c-component. Graphs $G20$, $G23$, $G25$, $G26$ and $G27$ in Table~\ref{tab:unident} illustrate such situation. These are discussed further in Section~\ref{subsec:unidentsemiMarkov}.

\subsubsection{Front-door adjustment}
\label{subsec:front}
In case a bi-directed edge between the sensitive attribute $A$ and the outcome $Y$ exists, all the above approaches will fail. However, $P(y_a)$ can still be measured using another criterion called the front-door criterion. The graph $G19$ satisfies this criterion. In fact, the back-door criterion cannot be used because of the unobserved confounder (impossible to control for) however, due to the presence of the mediators $M$, the front-door criterion can be applied to identify the causal effect as follows:
    \begin{equation}
        \label{eq:front}
	    P(y_a) = \sum_{m} \;P(m|a)\;\sum_{a'}P(y|m,a')\;P(a') 
    \end{equation}
More generally, the front-door adjustment can be applied if the the following conditions hold:
\begin{enumerate}
    \item all of the direct paths from $A$ to $Y$ pass through $M$. 
    \item there are no back-door paths from $A$ to $M$,
    \item all back-door paths from $M$ to $Y$ are blocked by $A$.
\end{enumerate}



Back-door and front-door adjustments are the main ingredients of the \textit{do-calculus}(Section~\ref{subsec:calculus}). 

\subsubsection{Instrumental variables}
\label{subsec:instru}
Consider graphs $G21$ and $G22$ in Table~\ref{tab:unident}. For non-parametric causal models, the causal effects $P(y_a)$ are not identifiable. However, for linear models, the causal effects for both graphs are identifiable using instrumental variables. A variable $I$ is said to be instrumental when it affects $A$, and only affects $Y$ by influencing $A$. That is, $I$ is an instrumental variable (relative to the pair $(A, Y)$ if (1) $I$ is independent of all variables (including unobserved variables) that have an influence on $Y$ that is not mediated by $A$ and (2) $I$ is not independent of $A$~\cite{pearl2009causality}. For graph $G21$, assuming a linear model, identifying the causal effect of $A$ on $Y$ is equivalent to identifying the coefficient $b = \frac{r_{IY}}{r_{IA}}$ on the edge: $A \rightarrow Y$. 
where: $r_{IY}$ is the slope of the regression line of $I$ on $Y$ and $r_{IA}$ is the slope of the regression line of $I$ on $A$. Hence, instrumental variables provide an efficient way to calculate the causal effect of $A$ on $Y$ without the need of controlling on the unobserved confounders. $I$ might be used to infer the causal effect $P(y_a)$ when some additional observed variable $Z$ can be used to control on such that: (1) $(I \not\!\perp\!\!\!\perp A |\;Z)_G$, (2) $(I \!\perp\!\!\!\perp Y |\;Z)_{G_{\overline A}}$. For example, in graph $G22$, $I$ is an instrument variable and by making back-door adjustments for $M$, we can identify $P(y_i)$ and $P(a_i)$. Since all the causal influence of $I$ on $Y$ must be channeled through $A$, we have:
     \begin{equation}
        \label{eq:instru}
	    P(y_i) = \sum_{a} P(y_a)\;P(a_i) 
    \end{equation}
    Thus, the causal effect of $A$ on $Y$ is identifiable  whenever Eq.~\ref{eq:instru} can be solved for $P(y_a)$ in terms of $P(y_i)$ and $P(a_i)$. 
\subsection{Non-identifiability: The hedge criterion}
\label{subsec:unidentsemiMarkov}
In some causal graphs neither the back-door, nor the front-door criteria are satisfied. The simplest graph where $P(y_a)$ cannot be calculated is the bow graph (graph $G20$ in Table~\ref{tab:unident}). The back-door criterion fails since the confounder variable is unobserved, while the front-door criterion fails since no intermediate variables between $A$ and $Y$ exist in the graph.

\begin{table}[!h]
\setlength\extrarowheight{3pt} 
    \begin{tabular}{|c|c|l|}
    \hline
         & \makecell{Causal graph\\} & \makecell{Hedge\\} \\
    \hline
       {\multirow{2}{*}{$G20$}}  & {\multirow{2}{*}{\includegraphics [scale=0.2]{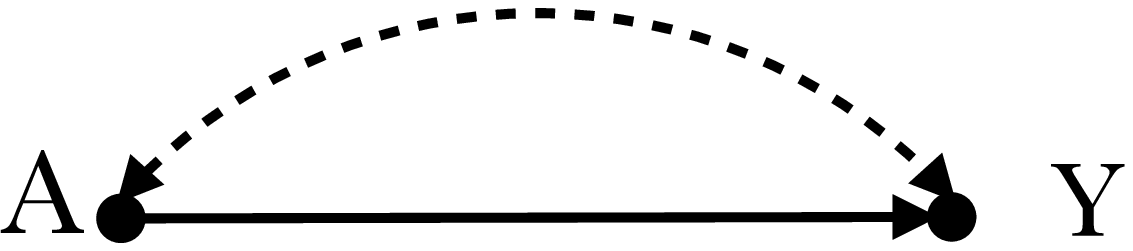}}} & $F:\{A,Y\}$\\
    & & \makecell[l]{$F':\{Y\}$\\\\}\\
    \hline 
    {\multirow{2}{*}{$G21^*$}}  & {\multirow{2}{*}{\includegraphics [scale=0.2]{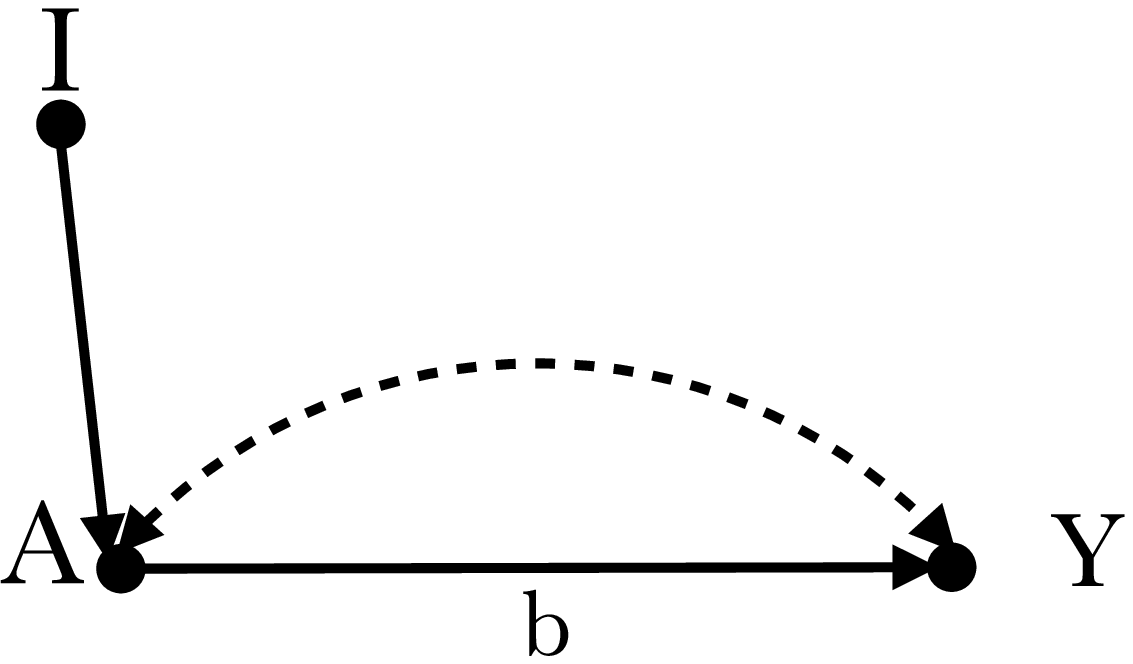}}} & $F:\{A,Y\}$\\
      & & \makecell[l]{$F':\{Y\}$\\\\}\\
      \hline
     {\multirow{3}{*}{$G22^*$}}  & {\multirow{3}{*}{\includegraphics [scale=0.2]{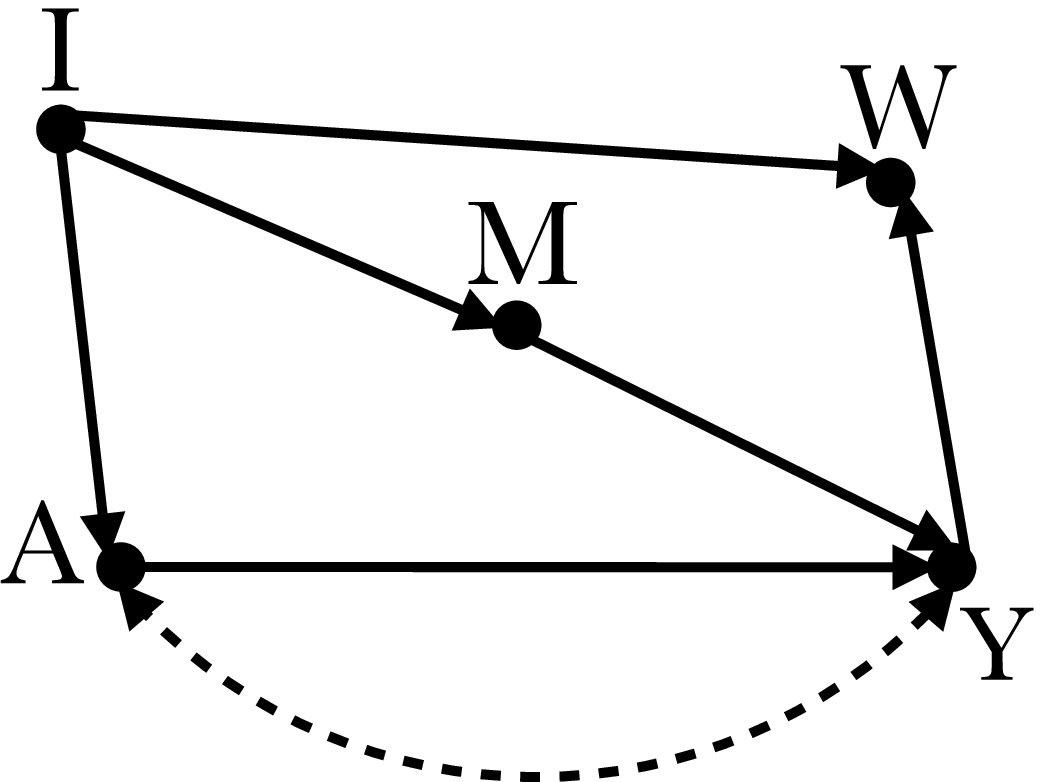}}} & $F:\{A,Y\}$\\
       & & \makecell[l]{$F':\{Y\}$\\\\}\\
        & & \\
    \hline
      {\multirow{3}{*}{$G23$}} & {\multirow{3}{*}{\includegraphics [scale=0.2]{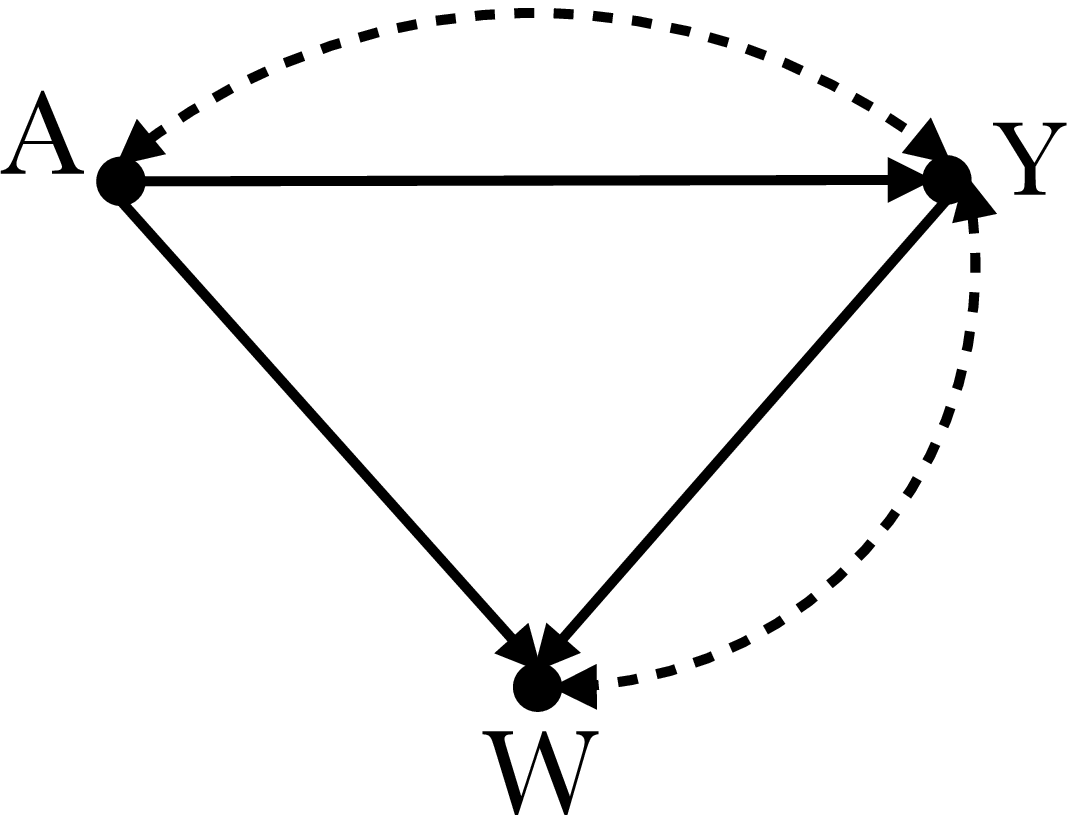}}} & $F:\{A,W,Y\}$\\
       & & \makecell[l]{$F':\{W,Y\}$\\\\}\\
        & & \\
    \hline
     {\multirow{3}{*}{$G24$}}  & {\multirow{3}{*}{\includegraphics [scale=0.2]{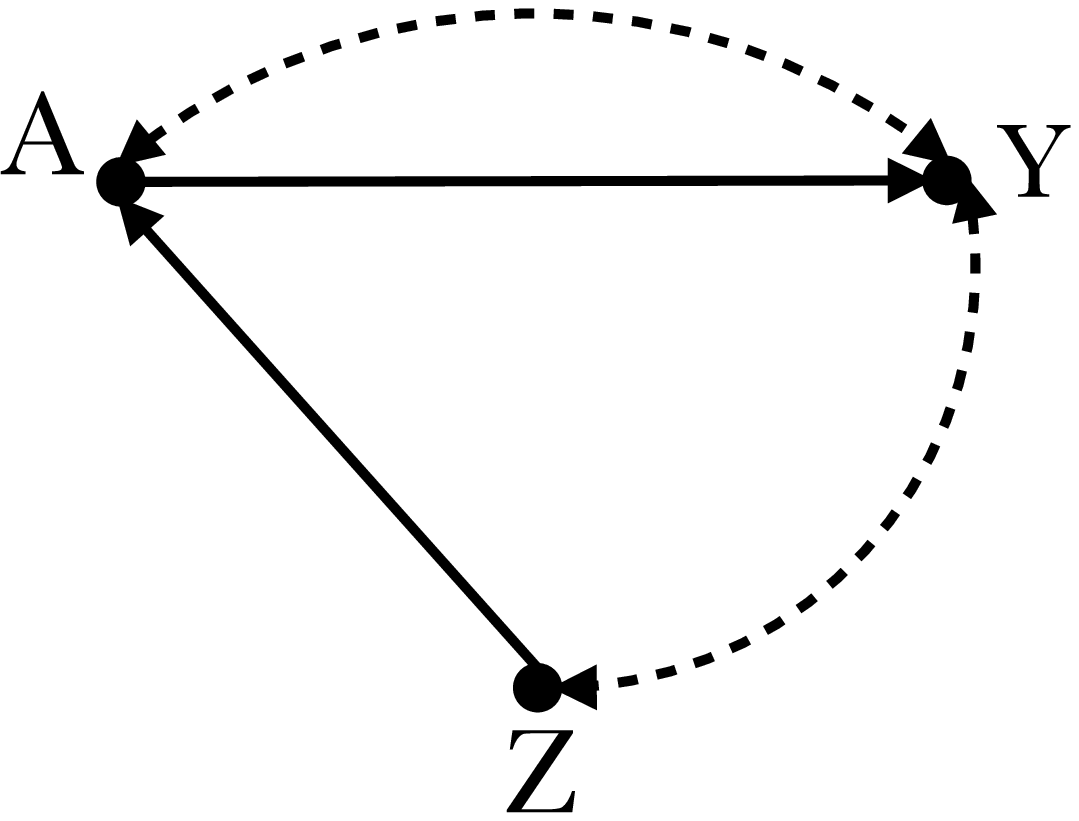}}} & $F:\{A,Z,Y\}$\\
        & & \makecell[l]{$F':\{Z,Y\}$\\\\}\\
         & & \\
    \hline
     {\multirow{2}{*}{$G25$}}  & {\multirow{2}{*}{\includegraphics [scale=0.2]{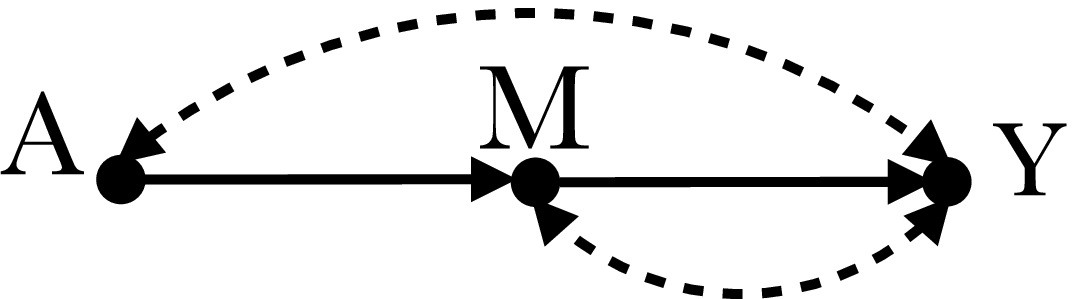}}} & $F:\{A,M,Y\}$\\
           & & \makecell[l]{$F':\{M,Y\}$\\\\}\\
    \hline
     {\multirow{3}{*}{$G26$}}  & {\multirow{3}{*}{\includegraphics [scale=0.2]{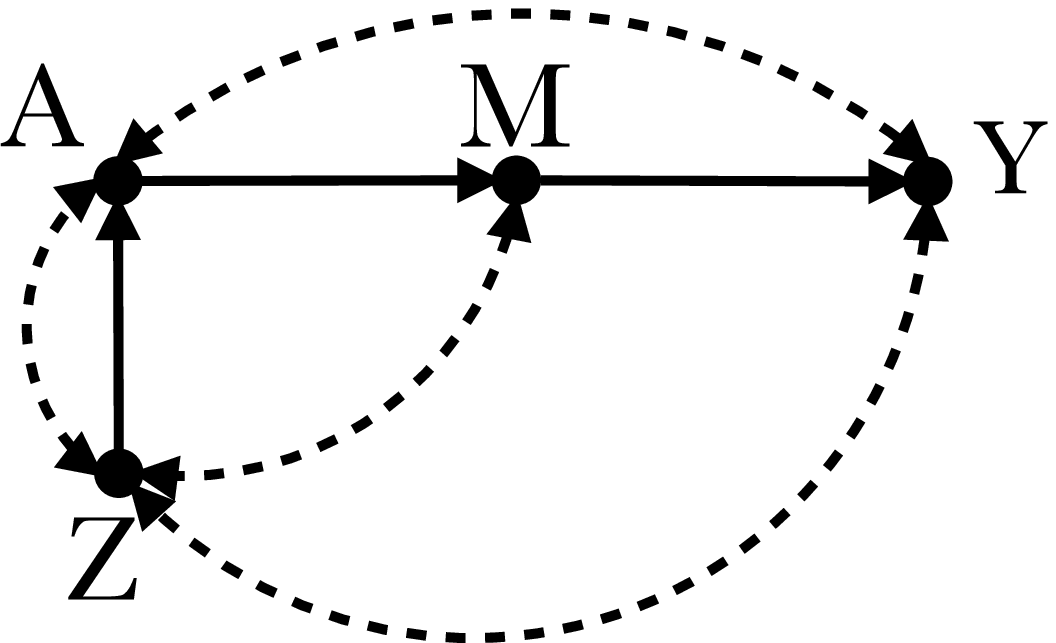}}} & $F:\{A,M,Z,Y\}$\\
         & & \makecell[l]{$F':\{M,Z,Y\}$\\\\}\\
          & & \\
    \hline
    {\multirow{3}{*}{$G27$}} & {\multirow{3}{*}{\includegraphics [scale=0.2]{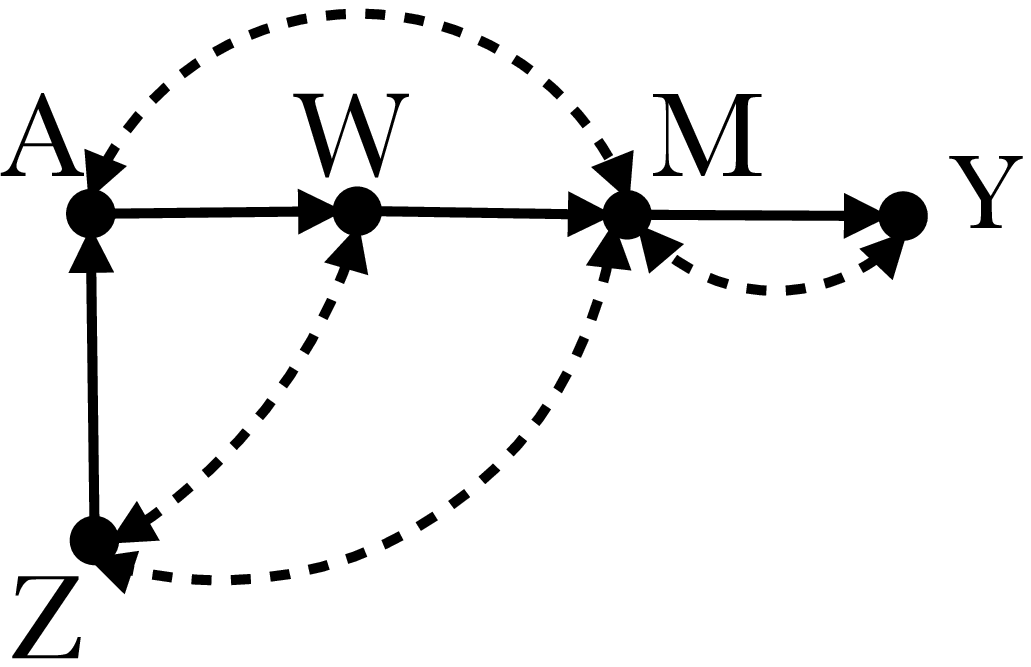}}} & $F:\{A,W,M,Z,Y\}$\\
           & & \makecell[l]{$F':\{W,M,Z,Y\}$\\\\}\\
             & & \\
    \hline
     {\multirow{2}{*}{$G28$}}  & {\multirow{2}{*}{\includegraphics [scale=0.2]{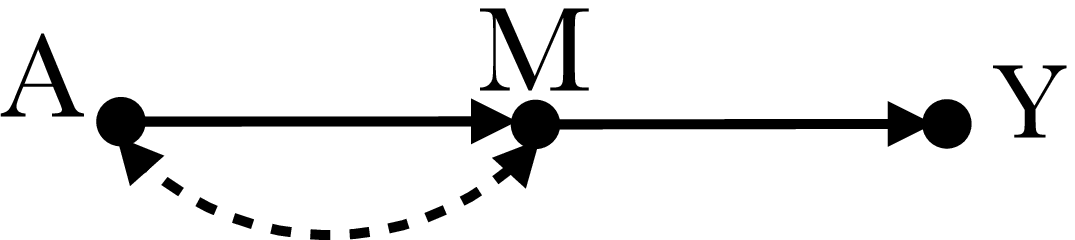}}} & $F:\{A,M\}$\\
       & & \makecell[l]{$F':\{M\}$\\\\}\\
    \hline
      {\multirow{2}{*}{$G29$}}  & {\multirow{2}{*}{\includegraphics [scale=0.2]{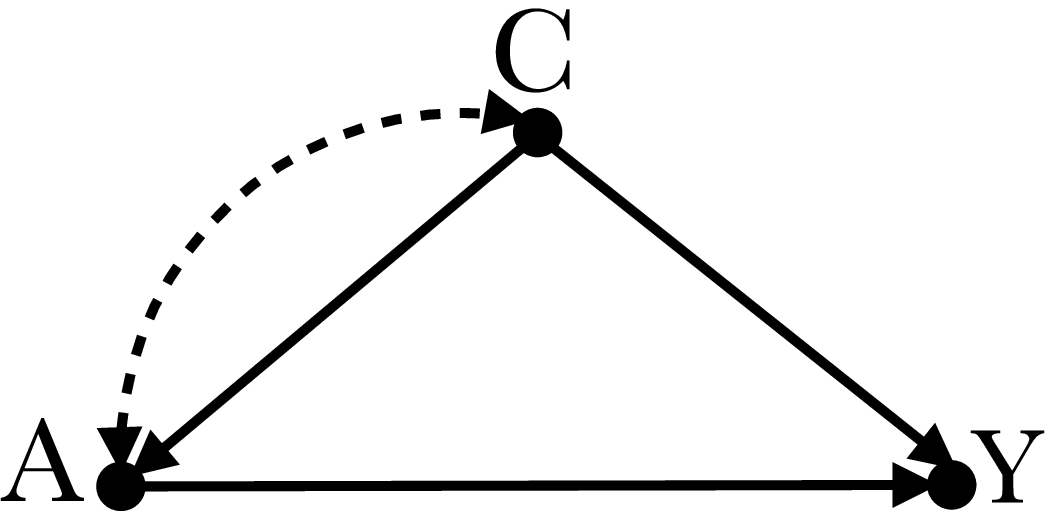}}} & $F:\{A,C\}$\\
       & & \makecell[l]{$F':\{C\}$\\\\}\\
    \hline
      {\multirow{2}{*}{$G30$}} & {\multirow{2}{*}{\includegraphics [scale=0.2]{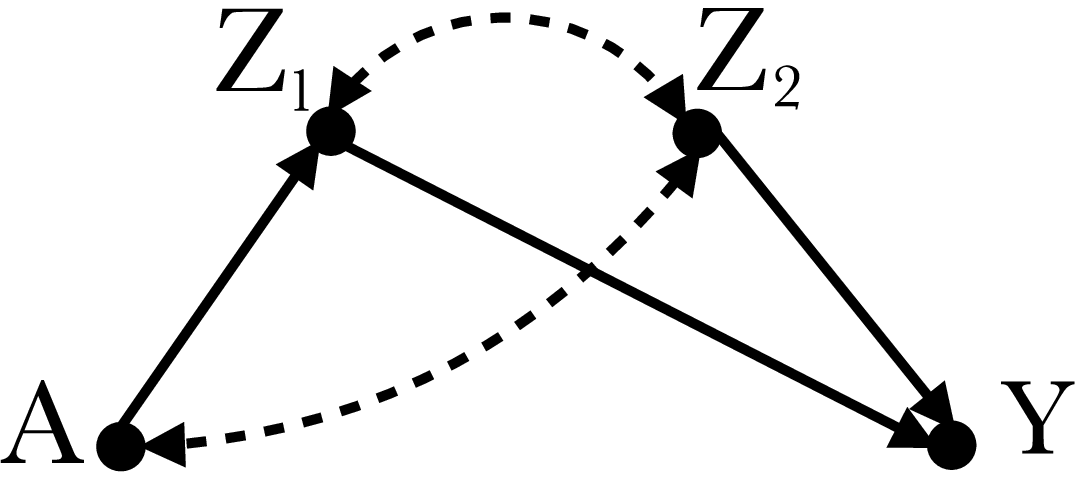}}} & $F:\{A,Z_1,Z_2\}$\\
      & & \makecell[l]{$F':\{Z_1,Z_2\}$\\\\}\\
    \hline
    \end{tabular}
    \caption{Non-identifiable semi-Markovian models. $^*$ indicates that the graph is identifiable in linear models.}
    \label{tab:unident}
\end{table}
Shpitser and Pearl~\cite{shpitser08} have  constructed a list of more generalized graphs inheriting the difficulty of the bow structure. Such graphs are called C-trees. A C-tree is a graph that is at the same time a tree\footnote{Notice that the direction of the arrows between nodes is reversed compared to the usual tree structure.} and a c-component (defined earlier in Section~\ref{subsec:Markov}). A tree is a graph such that each vertex (variable) has at most one child, and only one vertex (called the root) has no children. Graphs $G20-25$ are examples of Y-rooted C-trees (C-trees having as root the variable $Y$). Shpitser and Pearl proved that the causal effect of a Y-rooted C-tree is always unidentifiable (Theorem~12 in~\cite{shpitser08}). More generally, they demonstrated that all the unidentifiable cases of the causal effect $P(y_a)$ boil down to a general graphical structure called: \textit{hedge} which is defined as follows~\cite{shpitser08}:\newline
$F$ and $F'$ (sub-graphs in $G$) form a hedge for $P(y_a)$ if:
\begin{itemize}
    \item $F$ and $F'$ are \textbf{R}-rooted C-forests in $G$ (\textbf{R}$\in An(\mathbf{Y})_{G_{\overline{\mathbf{A}}}}$)
    \item $F'$ is a sub-graph of $F$ 
    \item $A$ only occurs in $F$ 
\end{itemize}
where a C-forest is a graph G which is both a c-component and a forest. And a forest is a graph G such that each vertex has at most one child. Note that any Y-rooted C-tree and its root node $Y$ form a hedge.

Table~\ref{tab:unident} presents other patterns where $P(y_a)$ is not identifiable due to the existence of a hedge structure in the graph. The graphs ($G26-30$) possess multiple c-components by contrast to $G20-25$. So, under such situations, the identifiability of the causal graph turns into the identifiability of each one of the c-components constituting this graph. In other words, each c-component is examined for a potential presence (or absence) of a hedge structure. It is sufficient to discover a hedge structure in one of the c-components to conclude that the whole causal graph is not identifiable (Theorem~19~\cite{shpitser08}). For example, the graph $G30$ is composed of two maximal c-components: $S1=\{A,Z_1,Z_2\}$ and $S2=\{Y\}$. Starting by the former, it is easy to recover a hedge structure: $F=\{A,Z_1,Z_2\}$ and $F'=\{Z_1,Z_2\}$ leading to the unidentifiability of the whole causal graph. Table~\ref{tab:unident} shows the sets $F$ and $F'$ for each graph.

Apart from measuring the causal effect for an identifiable causal graph, the \textbf{ID} algorithm~\cite{shpitser08} tells why $P(y_a)$ is not identifiable. That is, in case a particular graph is not identifiable, the algorithm returns the two sub-graphs $F$ and $F'$ that form a hedge in that graph.  

\section{Identification of counterfactual effects}
\label{sec:identctf}
While causal effects (Section~\ref{sec:identcausaleffects}) interpret the effect of actions as downward flow, counterfactual effects require more complex reasoning. Basically, counterfactual effects measure fairness based on multiple worlds: the actual world and other hypothetical (or counterfactual) worlds. The actual world is represented by a causal model $M$ in its actual (normal) state without any interventions, while the counterfactual worlds are represented by sub-models: $M_a$ where the intervention $do(a)$ forces the actual state to change to an alternative state. 

Note that in Markovian, as well as semi-Markovian models, if all parameters of the causal model are known (including $P(\mathbf{u})$), any counterfactual is identifiable and can be computed using the three steps abduction, action, and prediction (Theorem 7.1.7 in~\cite{pearl2009causality}). However, this method is usually infeasible in real-world scenarios due to the lack of the complete knowledge of the causal model (more specifically the knowledge of the background variables $U$). 

By contrast to causal effects, the calculation of counterfactual effects cannot only rely on the observational data summarized by $P(v)$ and the consensual causal graph $G$. In fact, counterfactual effects need, in addition, as input a set of experiments denoted as $P_*$ (called also: interventional probabilities). These experiments are formally defined as: $P_{*}= \{P_{\mathbf{x}}|\mathbf{X} \subseteq \mathbf{V},\;\mathbf{x}\;\text{\em x value assignment of }\; \mathbf{X}\}$ and are possible to perform in principle on a given causal model. Thus, the identifiability of counterfactuals depends on the identifiability of $P_{*}$ which in turn depends on $P(v)$ and $G$. Shpitser and Pearl~\cite{shpitser08} designed \textbf{ID*} and \textbf{IDC*} algorithms to evaluate counterfactuals given a causal graph and an observational data. In case these algorithms fail to uniquely measure a counterfactual quantity, the situation is referred to as an unidentifiable situation. 
\subsection{Counterfactual graph}
\label{subsec:ctfGraph}
\begin{figure}[!h]
\vspace{-3mm}
    \subfigure []  {%
    {\includegraphics [scale=0.22]{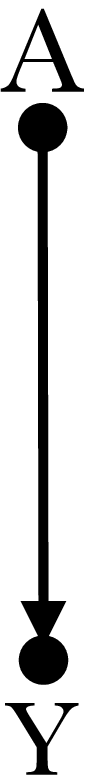} }
    \label{subfig:firing_ctf_a}}
    \qquad 
    \subfigure [] {%
    {\includegraphics[scale=0.25]{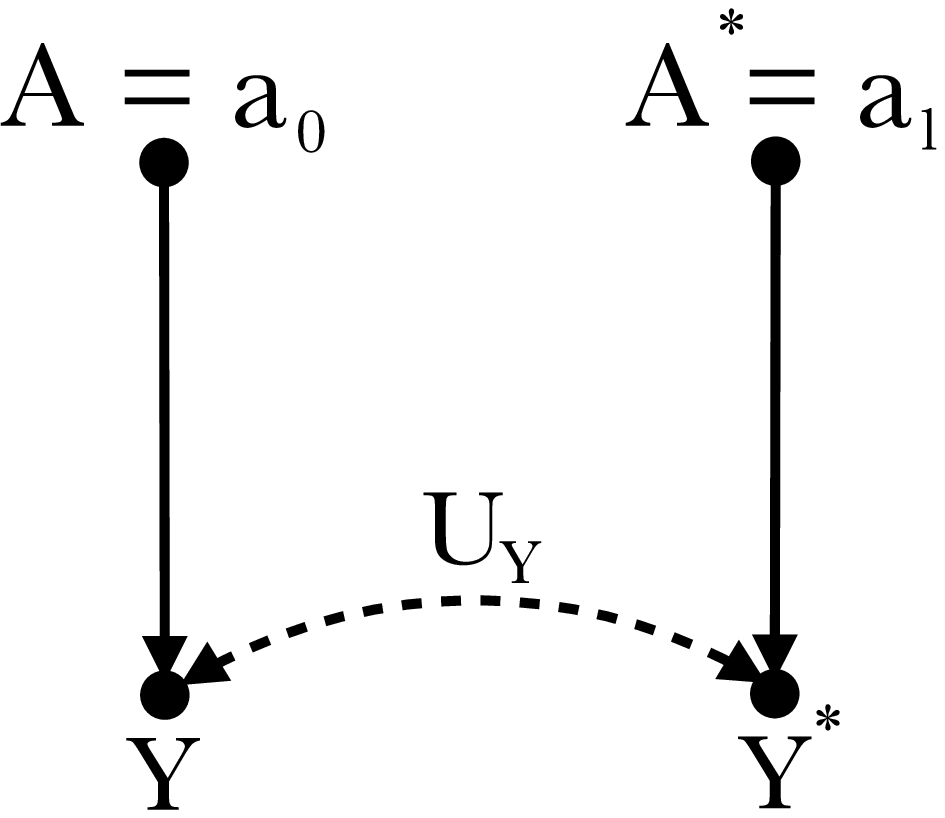} }
    \label{subfig:firing_ctf_b}}
    \qquad 
    \subfigure [] {%
    {\includegraphics[scale=0.25]{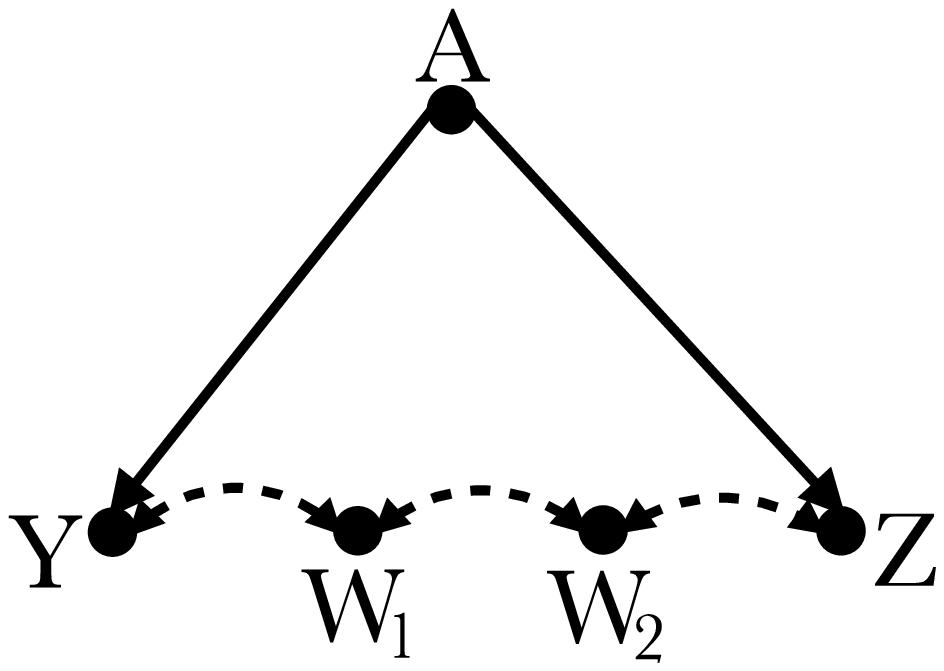} }
    \label{subfig:zigzag}}
      \caption{(a) A causal graph for the firing example (b) A corresponding counterfactual graph for the query $P(Y^*_{a*=a_1} |\; A=a_0$)(c) Zig-zag pattern.}
    \label{fig:firing_ctf}
    \vspace{-3mm}
\end{figure}

Given a causal graph $G$ of a Markovian model and a counterfactual expression $\gamma = v_{a}|e$ with $e$ some arbitrary set of evidence, measuring $P(\gamma)$ requires to construct a counterfactual graph which combines parallel worlds. Every world is represented by a counterfactual sub-model $M_{a}$. 
For example, Figure~\ref{fig:firing_ctf} shows a causal graph for the firing example (Figure~\ref{subfig:firing_ctf_a}) along with its corresponding counterfactual graph (Figure~\ref{subfig:firing_ctf_b}). Thus, Figure~\ref{subfig:firing_ctf_b} combines two worlds: the actual world where the teacher has actually $A = a_0$ and the counterfactual world where the same teacher is assigned $A^*\footnote{The subscript $^*$ is added to nodes belonging to the counterfactual world for graph legibility. } = a_1$. As shown in the figure, the two worlds share the same unobserved background variable: $U_Y$ that highlights the interaction between these worlds. Note that no bi-directed edges are connected to the node $A^*=a_1$. The reason for that is that the intervention $do(a^*=a_1)$ removes all the incoming arrows to $A^*$.
Thus, in order to calculate the counterfactual expression $P(Y^*_{a*=a_1} |\; A=a_0$) of the simple Markovian graph in Figure~\ref{subfig:firing_ctf_a}, we need to construct the semi-Markovian graph in Figure~\ref{subfig:firing_ctf_b}. The \textbf{make-cg} algorithm~\cite{shpitser08} automates this procedure. Basically, \textbf{make-cg} algorithm starts by combining the two causal graphs (actual and counterfactual) and makes them share the same background variable \textbf{U} (as shown in Figure~\ref{subfig:firing_ctf_b}). Then, it discards the duplicated endogenous nodes which are not affected by $do(a)$.

One typical unidentifiable counterfactual quantity is $P(y_{a'},y'_a)$ which is called the probability of necessity and sufficiency. The corresponding counterfactual graph is the W-graph that has the same structure as to Figure~\ref{subfig:firing_ctf_b}. This simple criterion can be generalized to the zig-zag graph (Figure~\ref{subfig:zigzag}) where the counterfactual $P(y_a,w_1,w_2,z_x')$ is not identifiable.

\subsection{C-component factorization}
\label{cptFact}
Now, in order to measure the counterfactual effects of a certain constructed counterfactual graph,
all the graphical criteria listed in Section~\ref{sec:identcausaleffects} hold. For instance, the c-component factorization approach can be used to decompose the counterfactual graph into a set of disjoint sub-graphs (or c-components). 
Thus, the joint distribution of all variables in the counterfactual graph can be factorized as the product of the conditional distribution of each c-component. Thus, if a certain causal effect is not identifiable in a particular c-component, the counterfactual quantity of the whole model is not identifiable as well.

\begin{figure}[!h]
\vspace{-3mm}
    \subfigure []  {%
    {\includegraphics [scale=0.2]{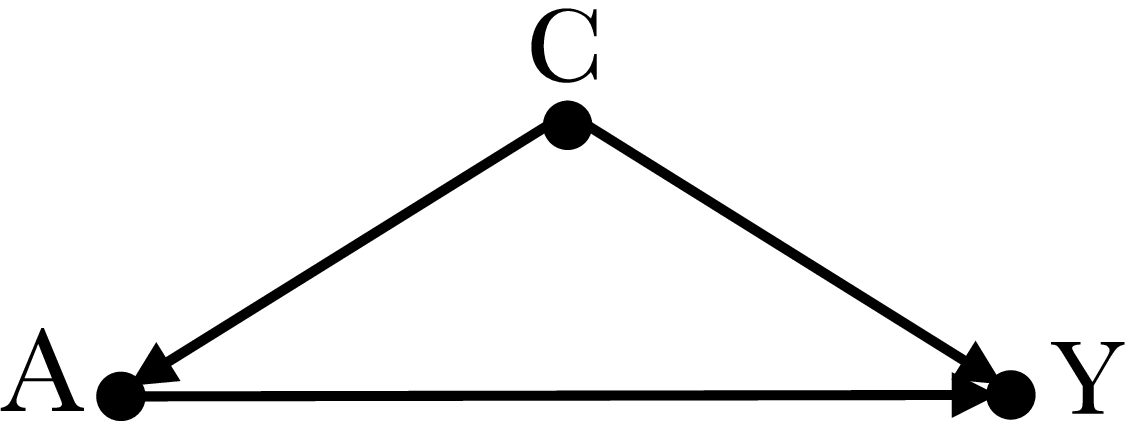} }
    \label{subfig:firing2a}}
    \quad \quad
    \subfigure [] {%
    {\includegraphics[scale=0.2]{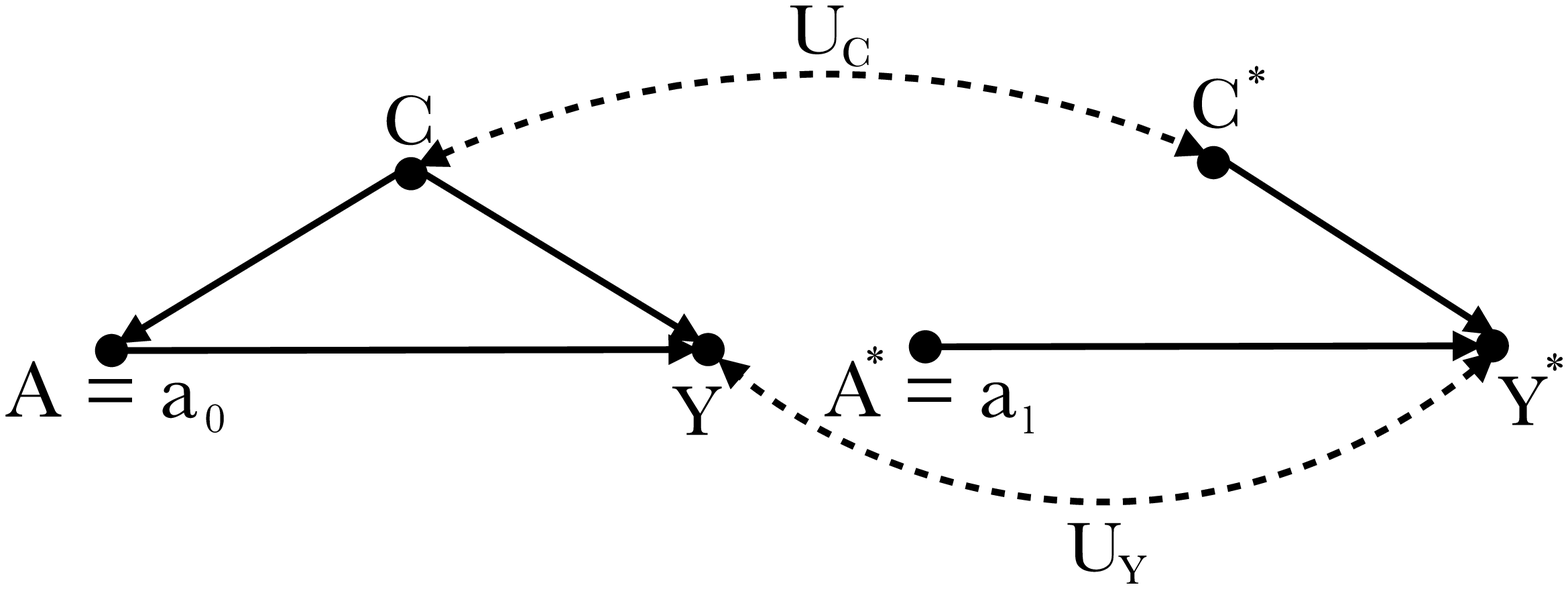} }
    \label{subfig:firing2b}}
    \\
    \subfigure [] {%
    {\includegraphics[scale=0.2]{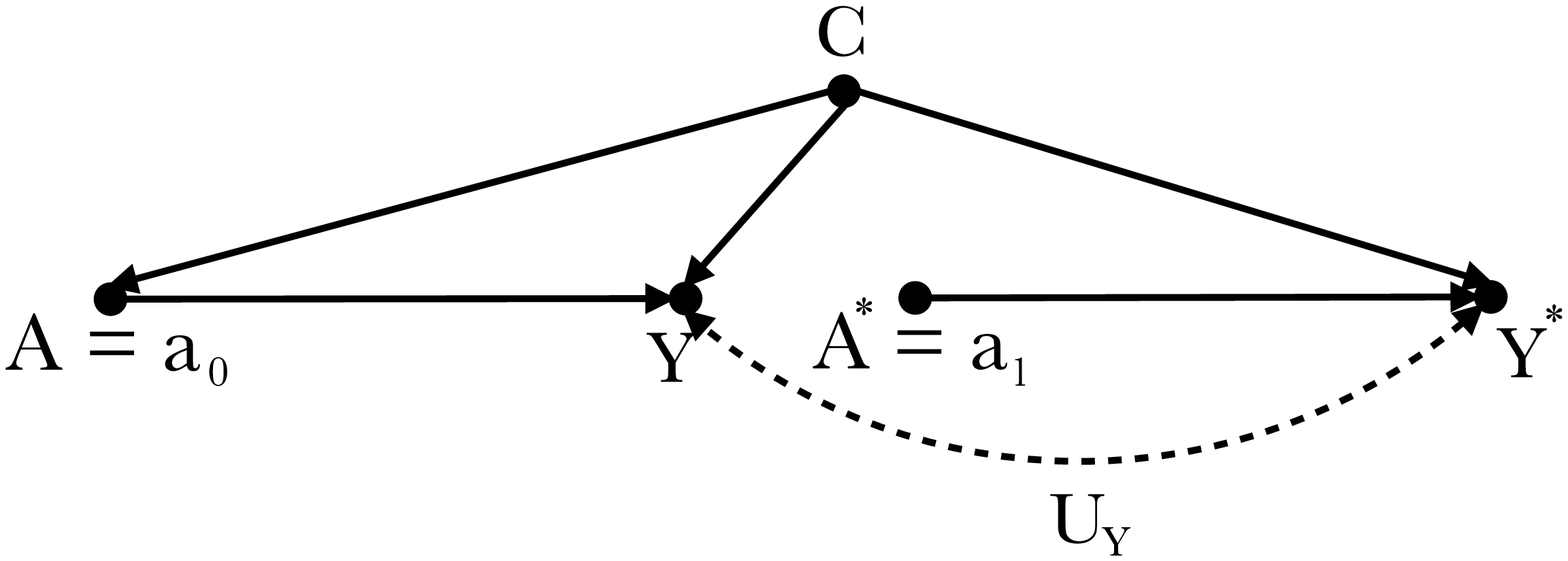} }
    \label{subfig:firing2c}}
      \caption{(a) Original causal graph for the firing example (b) Parallel worlds graph for $P(y_{a_1}|a_0)$ (c) Counterfactual graph for $P(y_{a_1}|a_0)$.}
    \label{fig:firing2_ctf}
    \vspace{-3mm}
\end{figure}

To illustrate how counterfactual quantities are measured, consider another simple scenario for the firing example. As shown in Figure~\ref{subfig:firing2a}, we assume now that the confounder variable (location of school) is observable. Consider the counterfactual query: $P(y_{a_1}|a_0)$ which reads the probability of firing a teacher who is assigned a class with a high initial level of students ($a_0$) had she been assigned a class with a low initial level of students ($a_1$). Figure~\ref{subfig:firing2b} shows the two parallel-worlds graph\footnote{This graph is called \textit{twin network} graph since it includes only two hypothetical graphs~\cite{shpitser08}.} for the query while Figure~\ref{subfig:firing2c} presents the final constructed counterfactual graph using \textbf{make-cg} algorithm. Note that in Figure~\ref{subfig:firing2c}, $C$ and $C^*$ are merged as a single node $C$ (by applying Lemma 24~\cite{shpitser08}). The main reason for that is that these nodes are not descendants of $A$. Then, $C$ inherits all the children of both nodes $C$ (the old node in the previous graph) and $C^*$. Finally, $U_C$ is omitted since any unobserved variable that possesses a single child should be removed~\cite{shpitser08}. 

Now, having constructed the counterfactual graph for the counterfactual expression $P(y_{a_1}|a_0)$, we can turn to the identifiability of this expression. Note that the obtained counterfactual graph (Figure~\ref{subfig:firing2c}) has three c-components: $\{C\},\{A\},\{Y,Y^*_{a_1}\}$ thus, applying algorithm $\mathbf{IDC^*}$~\cite{shpitser08} results in:
\begin{align}
	P(y_{a_1}|a_0) & = \frac{\sum_{y,c}\;Q(c)\;Q(a_0)\;Q(y,y_{a_1})}{P(a_0)}
\end{align}
where $Q(\mathbf{v}) = P(\mathbf{v}|pa(\mathbf{V}))$ in the counterfactual graph.

Hence, 
\begin{align}
	P(y_{a_1}|a_0) & = \frac{\sum_{y,c}\;P(c)\;P(a_0|c)\;P(y,y_{a_1}|c)}{P(a_0)} \nonumber\\
	& = \frac{\sum_{c}\;P(c)\;P(a_0|c)\;P(y_{a_1}|c)}{P(a_0)} \label{eq:cs2} \\
	& = \frac{\sum_{c}\;P(c)\;P(a_0|c)\;P(y|a_1,c)}{P(a_0)}\label{eq:cs3} \\ \nonumber
\end{align}

$y$ in Eq.~\ref{eq:cs2} is cancelled by summation while $P(y_{a_1}|c)$ in the same equation is transformed into $P(y|a_1,c)$ in Eq.~\ref{eq:cs3} using rule~(2) of the $do$-calculus (Section~\ref{subsec:identsemiMarkov}).

Table~\ref{tab:counterfactual} presents various examples of identifying the counterfactual quantities (column 3) of some causal graphs (first column) after obtaining their corresponding counterfactual graphs (column 2).   
\begin{table*}[t]
\setlength\extrarowheight{8pt} 
    \begin{tabular}{|c|c|c|l|}
    \hline
         & \makecell{Original Graph} & Counterfactual Graph & \makecell{$P(y_{a_1}|\;a_0)$} \\
    \hline
       \makecell{G31\\\\\\\\}& \includegraphics [scale=0.2]{./Graphics/FiringExample.eps}  & \includegraphics [scale=0.2]{./Graphics/ctfFiring.eps} & \makecell{unidentifiable if $y\neq y^*$\\\\\\\\} \\
    \hline
      \makecell{G32\\\\\\\\} &  \includegraphics [scale=0.2]{./Graphics/firing2a.eps}  & \includegraphics [scale=0.2]{./Graphics/firing2c.eps} & \makecell{$\frac{\sum_{c}\;P(c)\;P(a_0|c)\;P(y|a_1,c)}{P(a_0)}$\\\\\\\\\\} \\
    \hline
          \makecell{G33\\\\\\\\} &  \includegraphics [scale=0.2]{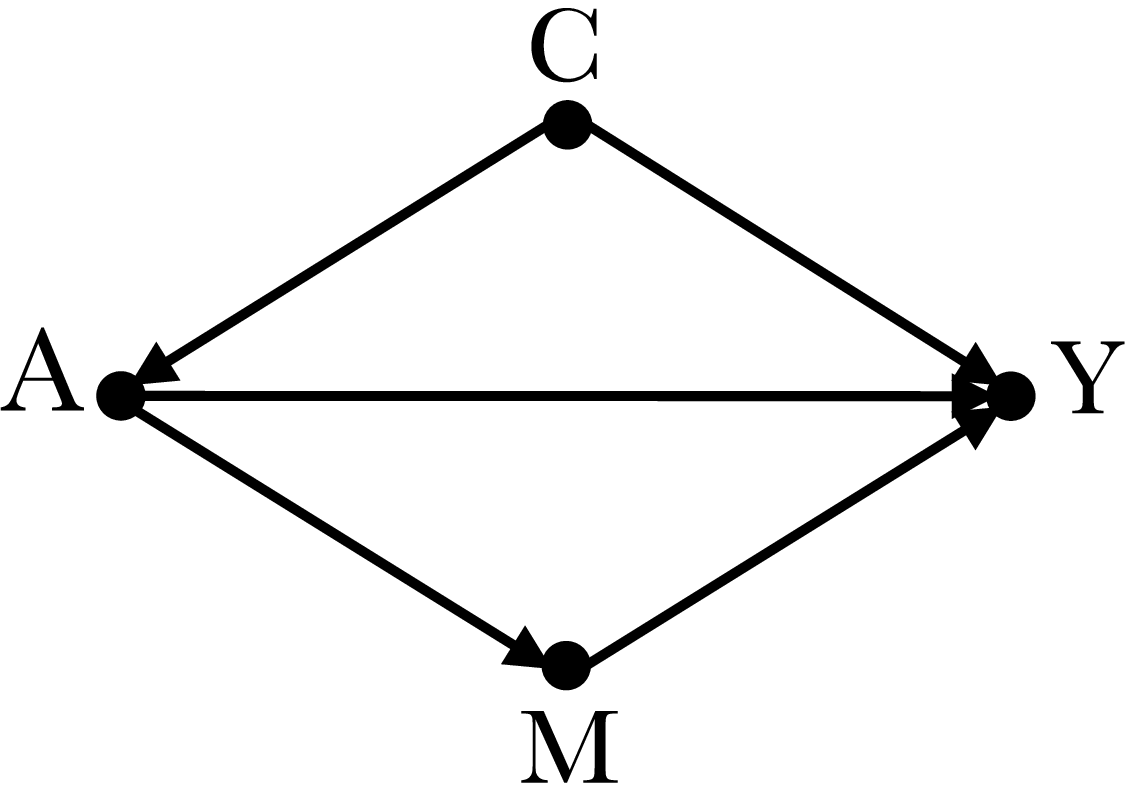}  & \includegraphics [scale=0.2]{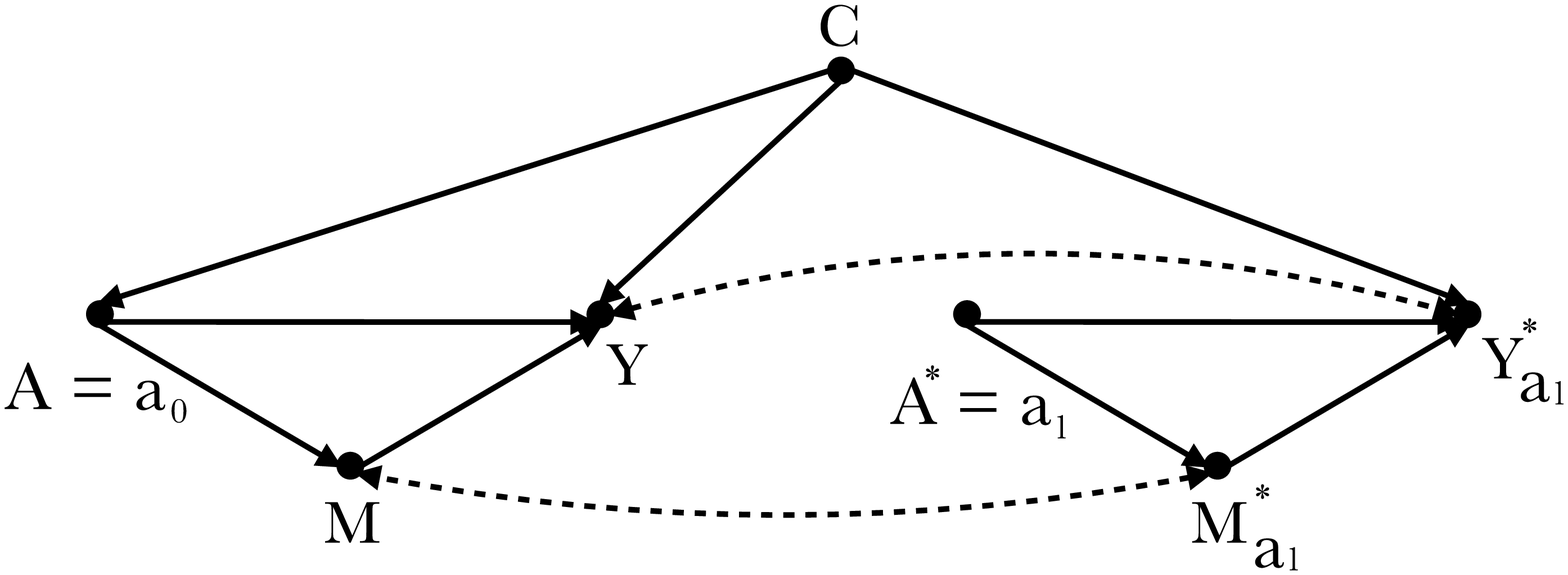} & \makecell{$\frac{\sum_{c,m}\;P(c)\;P(a_0|c)\;P(y|m,m',a_1,c)\;P(m'|\;a_0,c,a_1)}{P(a_0)}$\\\\\\\\\\} \\
    \hline
          \makecell{G34\\\\\\\\} &  \includegraphics [scale=0.2]{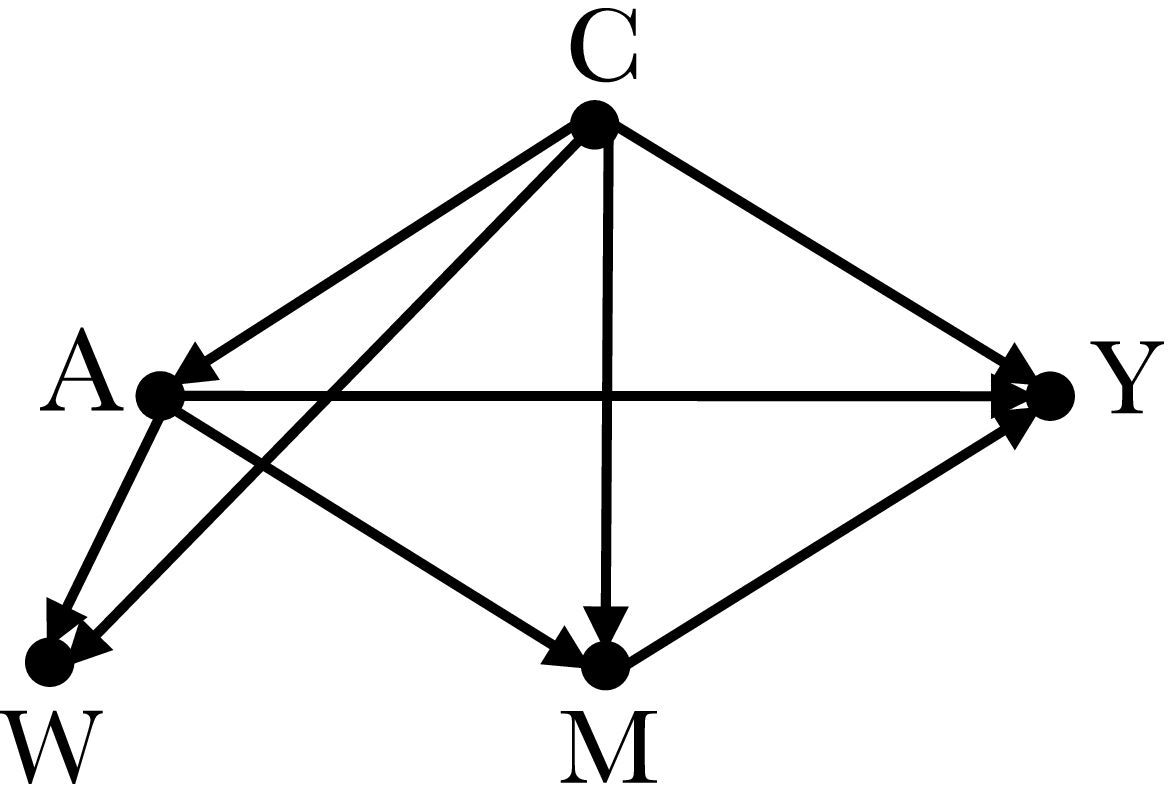}  & \includegraphics [scale=0.2]{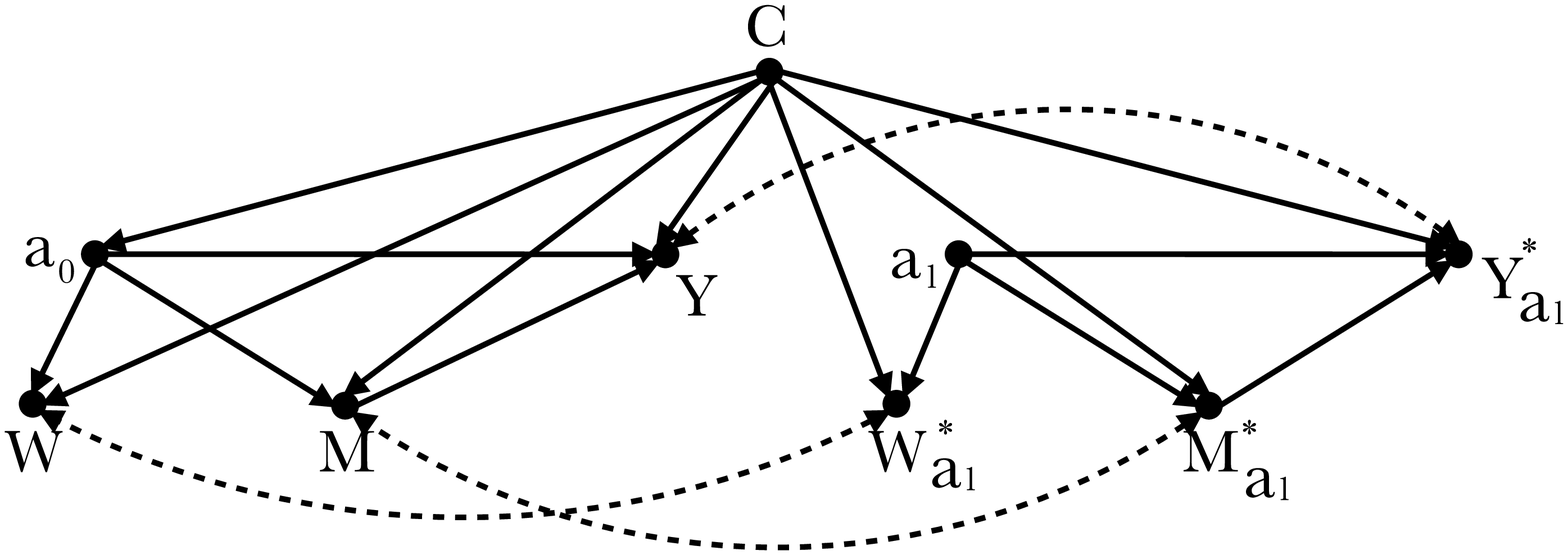} & \makecell{$\frac{\sum_{c}\;P(c)\;P(a_0|c)\;P(w|\;a_0,c)\;P(m'|\;a_1,c)\;P(y'|a_1,c,m')}{P(a_0)}$\\\\\\\\\\} \\
    \hline
    \end{tabular}
    \caption{Identifiability of counterfactual effects.}
    \label{tab:counterfactual}
\end{table*}

For example, $G33$ includes in addition to the confounder $C$ a mediator $M$. As shown in the corresponding counterfactual graph, the nodes $M$ and $M^*_{a_1}$ are not merged as they differ on their A-derived parents by contrast to the node $C$. Similarly, in $G34$, the pair of nodes $M$, $M^*_{a_1}$ and $W$, $W^*_{a_1}$ are not merged for the same reason.
\section{Identification of path-specific effects}
\label{sec:patheffect}
Identifying path-specific effects in a fairness context arises when one is interested in measuring the causal effect of the sensitive attribute $A$ on the outcome $Y$ only along certain pathways in the causal graph. That is, only the paths of interest are kept while all the other paths in the graph are excluded and not considered in the analysis. 

Direct and indirect effects are the simplest cases of path-specific effects. While the direct effect isolates the effect of $A$ on $Y$ along the direct path $A \rightarrow Y$, indirect effect considers the indirect causal paths between $A$ and $Y$ ($A \rightarrow \cdots \rightarrow Y$). 

A more general and complex case is when one wants to isolate the effect of $A$ on $Y$ along a specific group of paths. Such case is called path-specific effect. 
\subsection{Identification of direct and indirect effects}
\label{subsec:directindirect}
\begin{figure}[!h]
\vspace{-3mm}
    \subfigure []  {%
    {\includegraphics [scale=0.2]{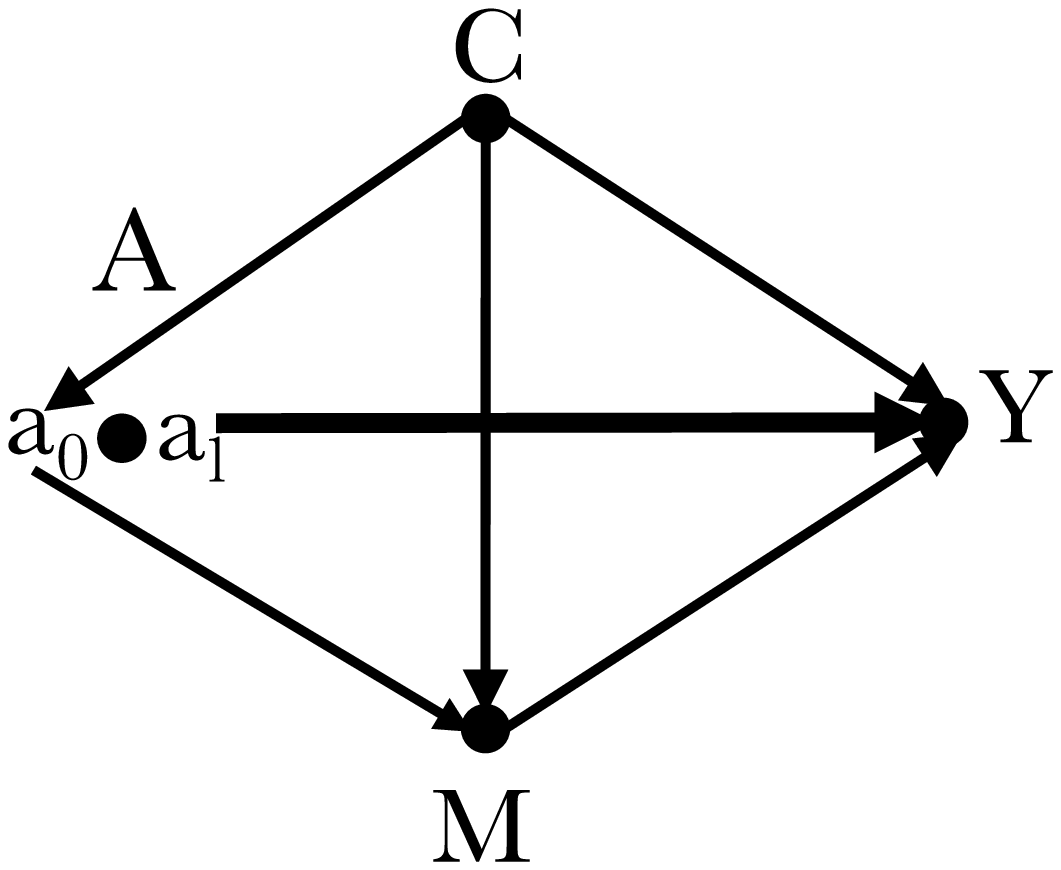} }
    \label{subfig:nde}}
    \quad 
    \subfigure [] {%
    {\includegraphics[scale=0.2]{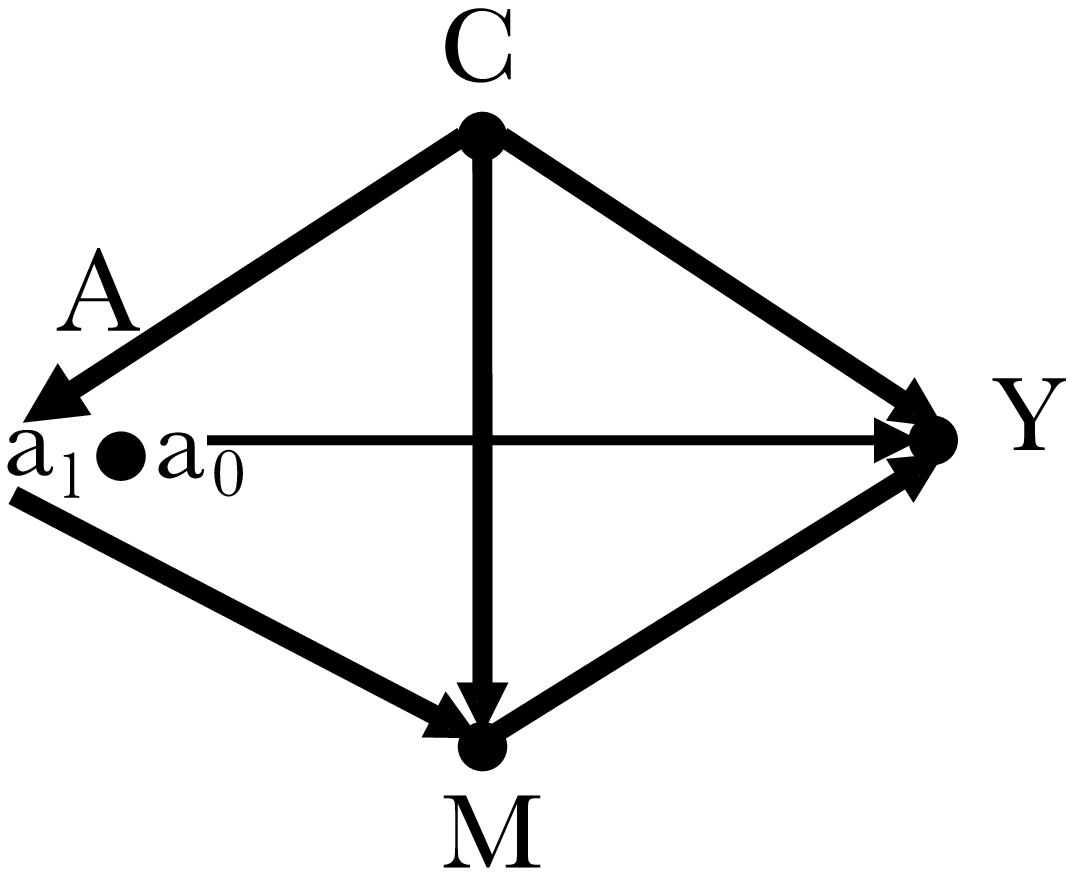} }
    \label{subfig:nie}}
    \quad 
    \subfigure [] {%
    {\includegraphics[scale=0.2]{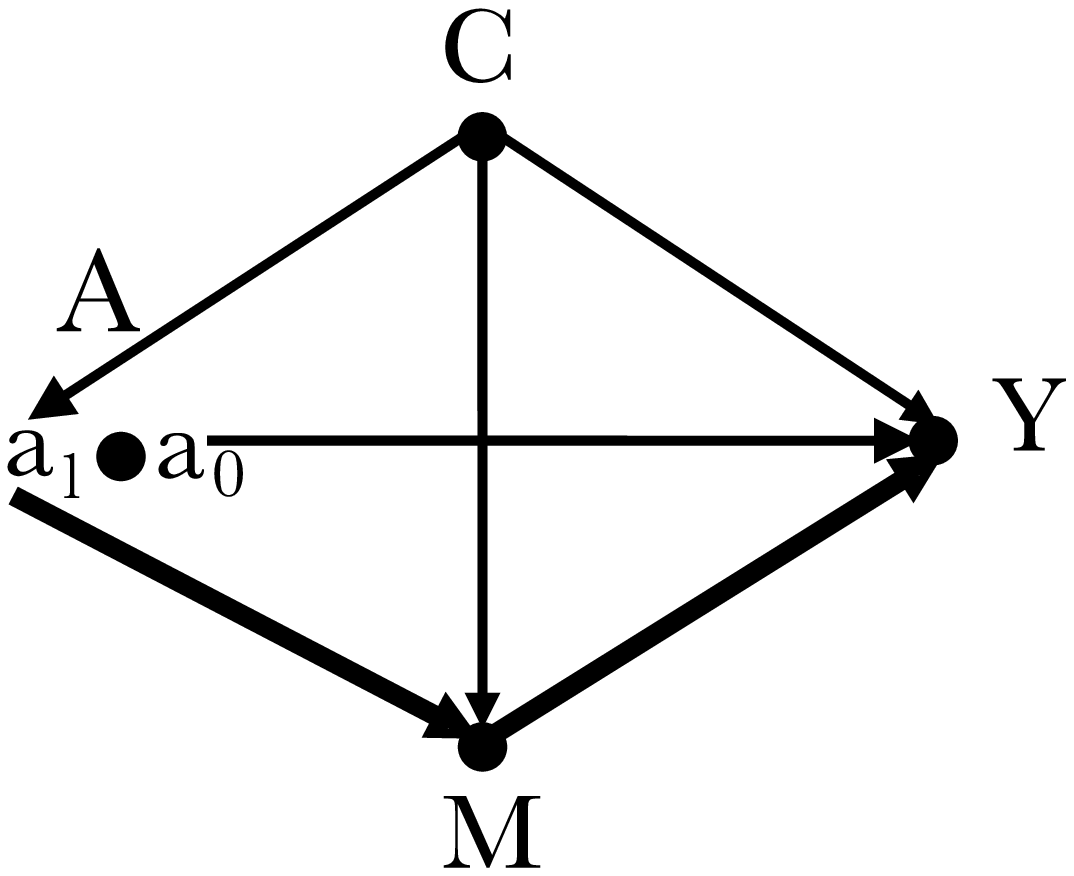} }
    \label{subfig:se}}
    \caption{Simple Markovian model that measures (a) \textit{NDE} (b) \textit{NIE} and (c) \textit{PSE} through the heavy edges.}
    \label{fig:nde_nie_pse}
    \vspace{-3mm}
\end{figure}

Average natural direct effect \textit{NDE} and Average natural indirect effect \textit{NIE} have been introduced by Pearl in~\cite{pearl01direct}. \textit{NDE} measures the total effect of $A$ on $Y$ that is not mediated by other variables $M$ in the causal model. In other words, \textit{NDE} evaluates the sensitivity of $Y$ to variations in $A$ while fixing all the other variables of the model. For example, the average natural direct effect in Figure~\ref{subfig:nde} is written as:
\begin{equation}
\label{eq:nde}
NDE_{a_1,a_0}(Y) = \textbf{E} [y_{a_1},M_{a_0}] - \textbf{E} [y_{a_0}]    
\end{equation}
where $\mathbf{E}[.]$ is the expectation of a random variable over all data inputs.

Considering the simple job hiring example (with $A=a_1$ = female, $A=a_0$ = male, $y$ = hiring, $M$= education level). Eq.~\ref{eq:nde} measures the expected change in male hiring ($\textbf{E} [y_{a_0}]$) had $A$ been $a_1$ (female), while mediators $\mathbf{M}$ are kept at the level they would take had $A$ been $a_0$ (e.g. male), in particular for the individuals $A=a_1$ (e.g. female).

\textit{NIE}, on the other hand, measures the effect of the mediator $M$ at levels $M_{a_0}$ and $M_{a_1}$ on $Y$ had $A$ been $a_0$ (Figure~\ref{subfig:nie}) and is defined as:
\begin{equation}
\label{eq:nie}
NIE_{a_1,a_0}(Y) = \textbf{E} [y_{a_0},M_{a_1}] - \textbf{E} [y_{a_0}]    
\end{equation}
Considering the same hiring example, Eq.~\ref{eq:nie} measures the expected change in male hiring $\textbf{E} [y_{a_0}]$, if males had equal education levels $(M=m)$ as those of females $(A=a_1)$.
Note that in the context of discrimination discovery, \textit{NIE} has been used under the assumption that $A$ has no parent node in the causal diagram (no spurious discrimination)~\cite{zhang2018fairness}.

In Markovian models, \textit{NDE} and \textit{NIE} are identifiable from observational data and can be calculated as follows~\cite{pearl01direct}:
\begin{equation}
\label{eq:identnde}
	NDE_{a_1,a_0}(Y) = \sum_{\mathbf{m}} \sum_{\mathbf{c}} \bigg(\mathbf{E}[Y|a_1,\mathbf{m}]-\mathbf{E}[Y|a_0,\mathbf{m}] \bigg)P(\mathbf{m}|a_0,\mathbf{c})P(\mathbf{c})
\end{equation}
\begin{equation}
\label{eq:identnie}
NIE_{a_1,a_0}(Y) = \sum_{\mathbf{m}} \sum_{\mathbf{c}} \mathbf{E}[Y|a_0,\mathbf{m}]\;\bigg (P(m|a_1,c)-P(m|a_0,c)\bigg )\;P(c)
\end{equation}
Where $\mathbf{M}$ is a set of mediator variables and $\mathbf{C}$ stands for any set satisfying the back-door criterion between the sensitive attribute $A$ and the outcome $Y$.
For instance, the graph $G37$ in Table~\ref{tab:nde_nie} illustrates a simple Markovian model for which the \textit{NDE} is given by Eq. \ref{eq:identnde} and the \textit{NIE} is given by Eq.~\ref{eq:identnie}. In case the effect of $A$ on $M$ is not confounded (graphs $G35$ and $G36$), \textit{NDE} is calculated using the following simplified equality:
\begin{equation}
    \label{eq:simplende}
    NDE_{a_1,a_0}(Y) = \sum_{\mathbf{m}} \bigg(\mathbf{E}[ Y|a_1,\mathbf{m}] -\mathbf{E}[Y|a_0,\mathbf{m}] \bigg ) P(\mathbf{m}|a_0)
\end{equation}
while \textit{NIE} is measured as follows:
\begin{equation}
    \label{eq:simplenie}
    NIE_{a_1,a_0}(Y) = \sum_{\mathbf{m}} \mathbf{E}[Y|a_0,\mathbf{m}] \bigg (P (m|\;a_1) - P (m|\;a_0)\bigg )
\end{equation}

\begin{table*}[!h]
\setlength\extrarowheight{8pt}
    \begin{tabular}{|c|c|l|}
    \hline
      &Causal graph   &  \makecell{Identifiability Formula} \\
    \hline
       \makecell{G35\\\\}& \includegraphics [scale=0.2]{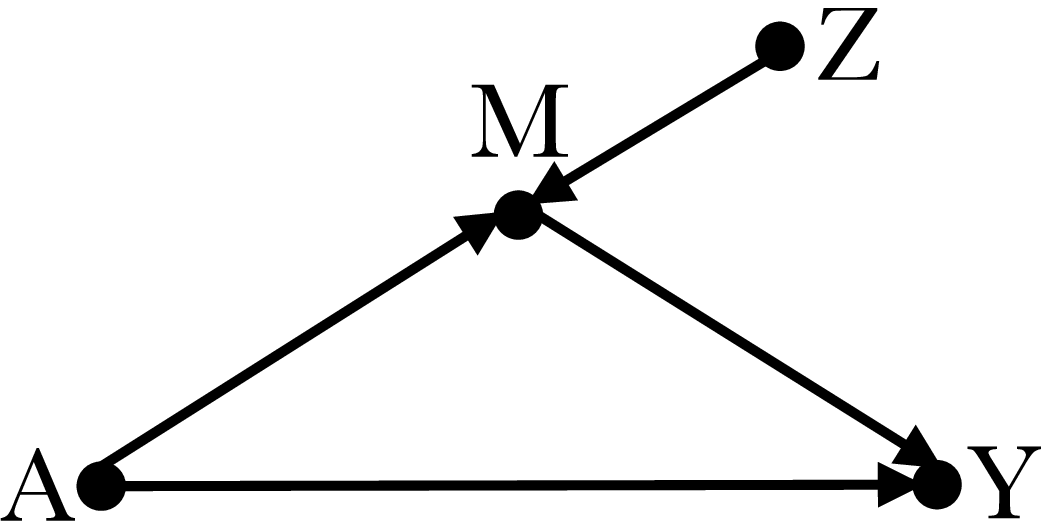} & \makecell[l]{$NDE_{a_1,a_0}(Y) = \sum_{\mathbf{m}} \bigg (\mathbf{E}[ Y|a_1,\mathbf{m}] -\mathbf{E}[Y|a_0,\mathbf{m}] \bigg ) P(\mathbf{m}|a_0)$\\} \\
    \cline{1-2}
     \makecell{G36\\\\} & \includegraphics [scale=0.2]{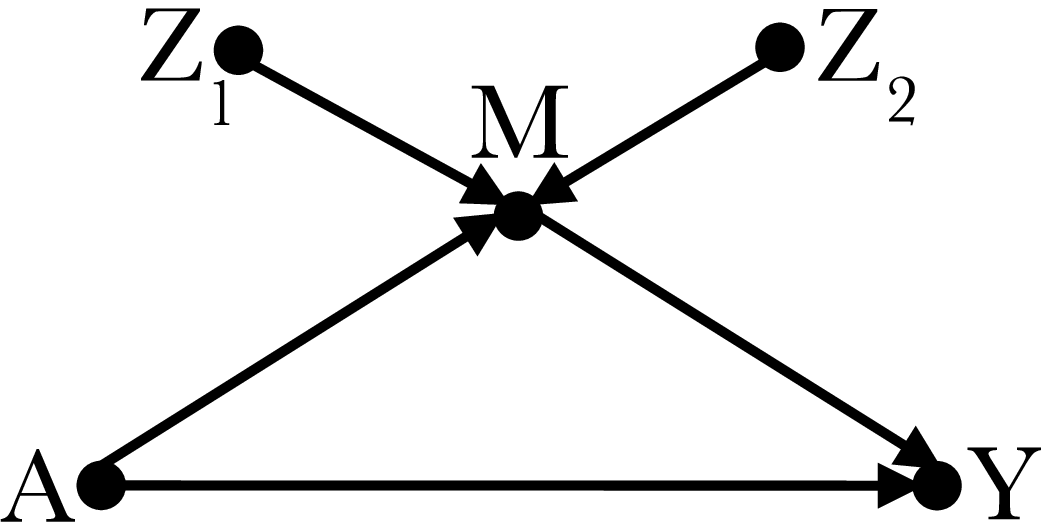}  & \makecell[l]{$NIE_{a_1,a_0}(Y) = \sum_{\mathbf{m}} \mathbf{E}[Y|a_0,\mathbf{m}] \bigg (P (m|\;a_1) - P (m|\;a_0)\bigg )$\\\\\\}\\
    \hline
          \makecell{{\multirow{2}{*}{G37}}} &  {\multirow{2}{*}{\includegraphics [scale=0.2]{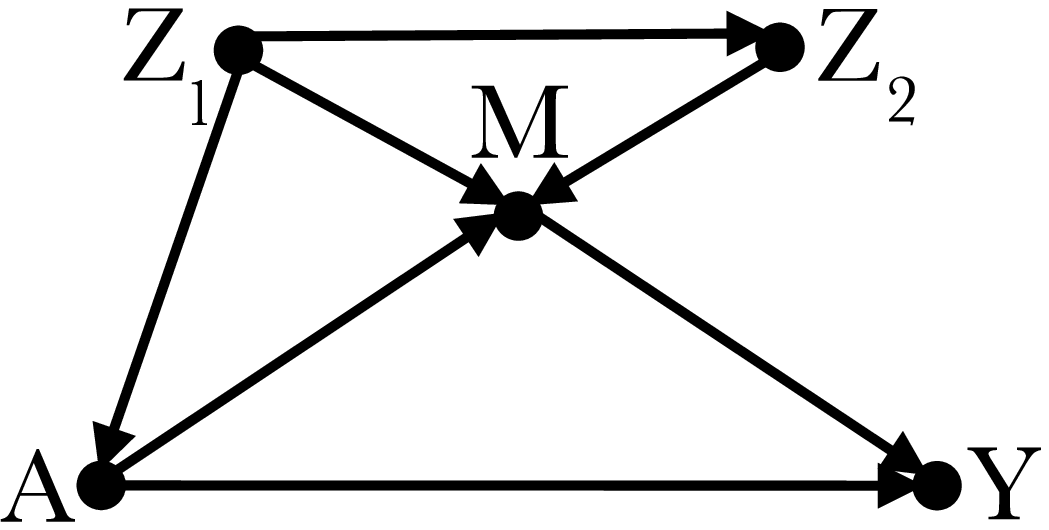}}}  & $NDE_{a_1,a_0}(Y) = \sum_{\mathbf{z_2}} \sum_{\mathbf{m}} \bigg(\mathbf{E}[Y|a_1,\mathbf{m}] -\mathbf{E}[Y|a_0,\mathbf{m}] \bigg)P(\mathbf{m}|a_0,\mathbf{z_2})P(\mathbf{z_2})$\\
    & & $NIE_{a_1,a_0}(Y) = \sum_{\mathbf{z_2}} \sum_{\mathbf{m}} \mathbf{E}[Y|a_0,\mathbf{m}] \bigg (P(m|a_1,z_2)-P(m|a_0,z_2)\bigg )\;P(z_2)$\\
        \hline
    \end{tabular}
    \caption{Identifiability of NDE and NIE.}
    \label{tab:nde_nie}
\end{table*}

In semi-Markovian models, \textit{NDE} and \textit{NIE} are not generally identifiable. However, if there exists a set $Z$ of covariates, non-descendants of $A$ or $M$, such that, for all values of $a$ and $m$ we have the following conditions (Theorem 2~\cite{pearl01direct}):
\[
    \text{(1)} \quad Y_{a_1 m} \!\perp\!\!\!\perp M_{a_0} |\;Z 
\]
\[
    \text{(2)} \quad P(Y_{a_1 m} = y |\; Z=z) \quad \text{is identifiable}
\]
\[
    \text{(3)} \quad P(M_{a_0} = m |\; Z=z) \quad \text{is identifiable}
\]

then, \textit{NDE} is identifiable and is calculated as follows:
\begin{equation}
    \label{eq:nde_semi}
    NDE_{a_1,a_0}(Y) = \sum_{m,z}\bigg (\mathbf{E}[Y_{a_1,m}|\;z] - \mathbf{E}[Y_{a_0,m}|\;z]\bigg )\;P(M_{a_0}=m|\;z)\;P(z) 
\end{equation}

Similarly, \textit{NIE} is identifiable in semi-Markovian models using observational data and is given by (Theorem~4~\cite{pearl01direct}):
\begin{equation}
    \label{eq:nie_semi} 
    NIE_{a_1,a_0}(Y) = \sum_{m,z} \mathbf{E}[Y_{a_0,m}|\;z]\bigg (P(M_{a_1}=m|\;z)-P(M_{a_0}=m|\;z) \bigg )P(z) 
\end{equation}
if the following expressions are satisfied for all values of $a$ and $m$:
\[
    \text{(1)} \quad Y_{a_0,m} \!\perp\!\!\!\perp M_{a_1} |\;Z 
\]
\[
    \text{(2)} \quad \mathbf{E} [Y_{a_0,m} |\; z] \quad \text{is identifiable}
\]
\[
    \text{(3)} \quad P(M_{a_1} = m |\; z) \quad \text{is identifiable}
\]
\[
    \text{(3)} \quad P(M_{a_0} = m |\; z) \quad \text{is identifiable}
\]

\subsection{Identification of path-specific effects}
\label{subsec:pathspecific}
One of the challenges of discrimination discovery is to distinguish between two types of indirect effects of the sensitive attribute $A$ on the outcome $Y$, namely: the indirect discrimination (unfair effect) and the explainable effect (fair effect). While the simple \textit{NIE} (explained in the previous section) considers all the indirect paths between $A$ and $Y$ regardless of whether they are fair or not, path-specific effect provides a more fine-grained way to consider effects along a selected subset of paths between $A$ and $Y$. In 
other words, path-specific effect makes the distinction between the two types of indirect effects possible. Figure~\ref{subfig:se} shows an example of a causal graph where only the heavy path ($A \rightarrow M \rightarrow Y$) is selected for effect analysis.
Given a path set $\pi$, the $\pi$-specific effect~\cite{pearl01direct} is defined as: 
\begin{equation}
\label{eq:PSE}
PSE^{\pi}_{a_1,a_0}(y) = P(y_{a_1 |_\pi, a_0 |_{\overline{\pi}}}) - P(y_{a_0})
\end{equation}
where $P(y_{a_1 |_\pi, a_0 |_{\bar{\pi}}})$ is the probability of $Y=y$ in the counterfactual situation where the effect of $A$ on $Y$ with the intervention ($A=a_1$) is transmitted along $\pi$, while the effect of $A$ on $Y$ without the intervention ($A=a_0$) is transmitted along paths not in $\pi$ (denoted by: $\bar{\pi}$).

The identifiability of $PSE^{\pi}_{a_1,a_0}(y)$ depends on the identifiability of the term $P(y_{a_1 |_\pi, a_0 |_{\bar{\pi}}})$. Avin et al.~\cite{avin2005identifiability} provided the necessary and sufficient condition for $P(y_{a_1 |_\pi, a_0 |_{\bar{\pi}}})$ to be identifiable in Markovian models, namely, the recanting witness criterion. 

\begin{figure}[!h]
\vspace{-3mm}
    \subfigure []  {%
    {\includegraphics [scale=0.25]{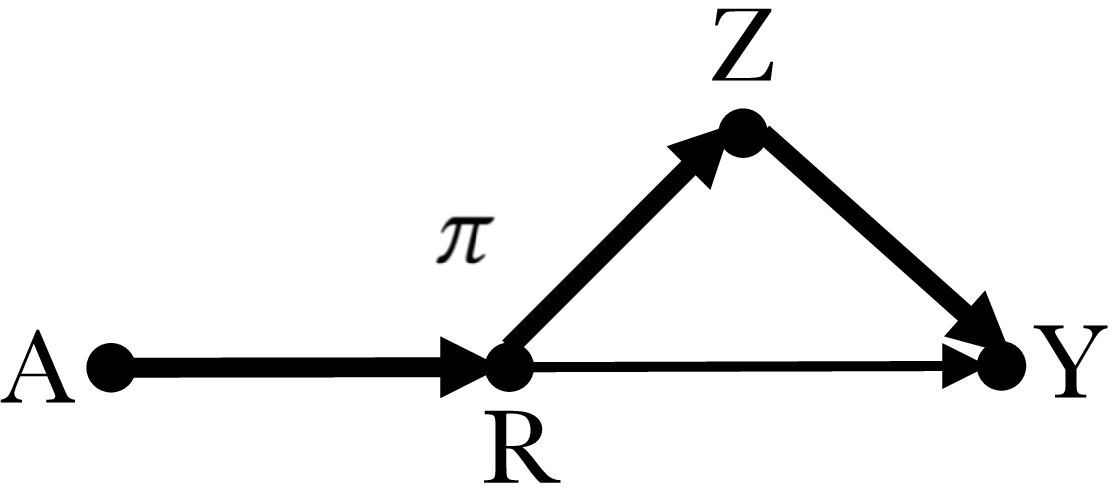} }
    \label{subfig:kite}}
    \qquad 
    \subfigure [] {%
    {\includegraphics[scale=0.25]{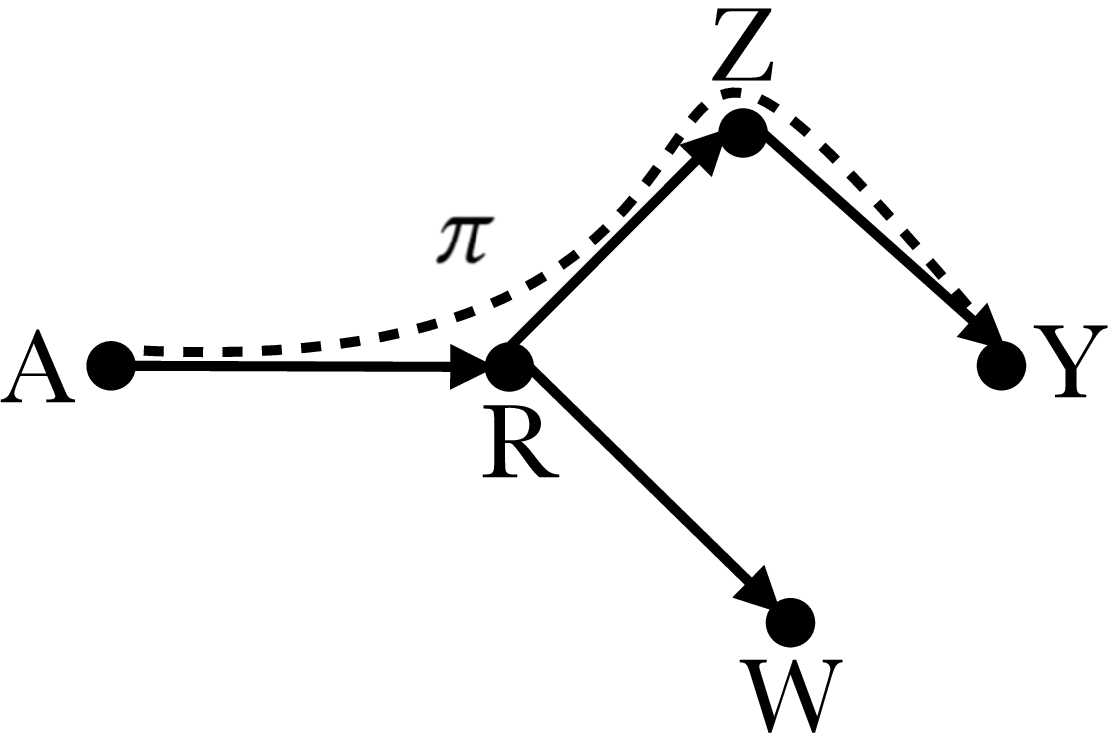} }
    \label{subfig:pse11}}
    \caption{(a) The recanting witness criterion satisfied (the ``kite'' pattern). (b) illustrates an example of identifiable $\pi$-specific effect graphs. $\pi$ path is represented by a heavy line.}
    \label{fig:nde_nie}
    \vspace{-3mm}
\end{figure}

Given a path $\pi$ in $G$ pointing from $A$ to $Y$, the recanting witness criterion is satisfied for $\pi$-specific effect if and only if there exists a variable $R$ (known as witness) in $G$ such that: (1) there exists a path from $A$ to $R$ in $\pi$, (2) there exists a path from $R$ to $Y$ in $\pi$, and (3) there exists another path from $R$ to $Y$ not in $\pi$. The graphical pattern of this criterion is called the ``kite'' pattern and is shown in Figure~\ref{subfig:kite}. In this graph, the witness variable is $R$ thus, $\pi= A \rightarrow R \rightarrow Z \rightarrow Y$ while $\overline{\pi}= A \rightarrow R \rightarrow Y$. 

Hence, if the recanting witness criterion is satisfied, $PSE^{\pi}_{a_1,a_0}(y)$ is not identifiable. Figure~\ref{subfig:kite} illustrates a simple example where $PSE^{\pi}_{a_1,a_0}(y)$ is identifiable as the recanting witness criterion is not satisfied. Under such setting, $P(y_{a_1 |_\pi, a_0 |_{\bar{\pi}}})$ can be measured by applying the following steps: 
\begin{enumerate}
    \item Express $P(y_{a_0})$ using the truncated factorization formula according to Eq. ~\ref{eq:trunc}.
    \item Segregate $Ch(A)$ other than Y ($Ch(A)\backslash Y$) into two sets: $\mathbf{S_1}$ and $\mathbf{S_2}$ (where: $ \mathbf{S_1} \cap \mathbf{S_2} = \emptyset $ ). The nodes $\in \mathbf{S_1}$ belong to edges in  $\pi$ while nodes $\in \mathbf{S_2}$ belong to edges not in $\pi$ or not included in any path from $A$ to $Y$.
    \item Replace values $a_0$ with $a_1$ for the terms corresponding to nodes in $\mathbf{S_1}$, and keep values $a_0$ unchanged for the terms corresponding to nodes in $\mathbf{S_2}$ (Theorem~2~\cite{shpitser2013counterfactual}).
\end{enumerate}

For example, in Figure~\ref{subfig:kite}, $S_1 = \{R\}$ and $S_2= \emptyset$. Thus, $P(y_{a_1 |_\pi, a_0 |_{\bar{\pi}}})$ is identifiable and is given by:
\[
    \sum_{{r,w,z}} P(r|\;a_1) P(z|\;r)\;P(w|\;r)\;P(y|\;z)
\]

Table~\ref{tab:pse} presents other examples of Markovian models where $\pi$-specific effect is identifiable. The formula of how the quantity: $P(y_{a_1 |_\pi, a_0 |_{\bar{\pi}}})$ is calculated for each example and is shown in the third column.

\begin{table}[!h]
    \begin{tabular}{|c|c|l|}
    \hline
      &Causal graph   &  \makecell{$P(y_{a_1 |_\pi, a_0 |_{\overline{\pi}}})$} \\
    \hline
       \makecell{G38\\\\\\\\}& \includegraphics [scale=0.2]{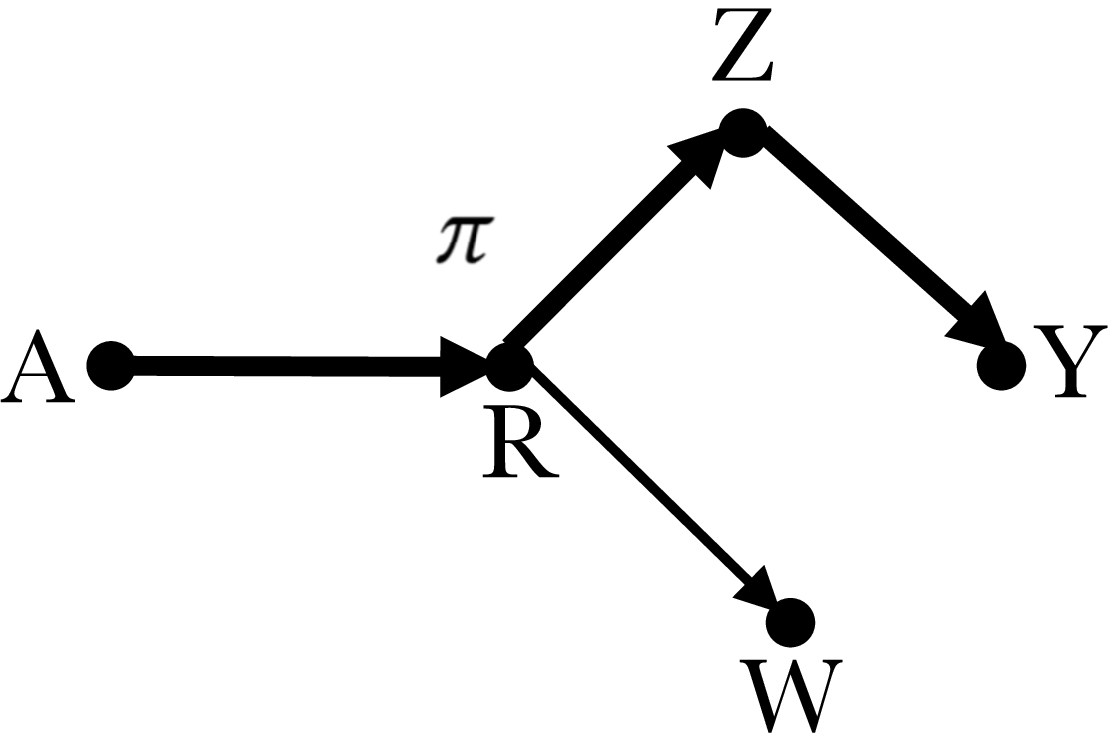} & \makecell[l]{$ \sum_{{r,w,z}} P(r|\;a_1) P(z|\;r)\;P(w|\;r)\;$\\$P(y|\;z)$} \\
    \hline
      \makecell{G39\\\\\\\\} & \includegraphics [scale=0.2]{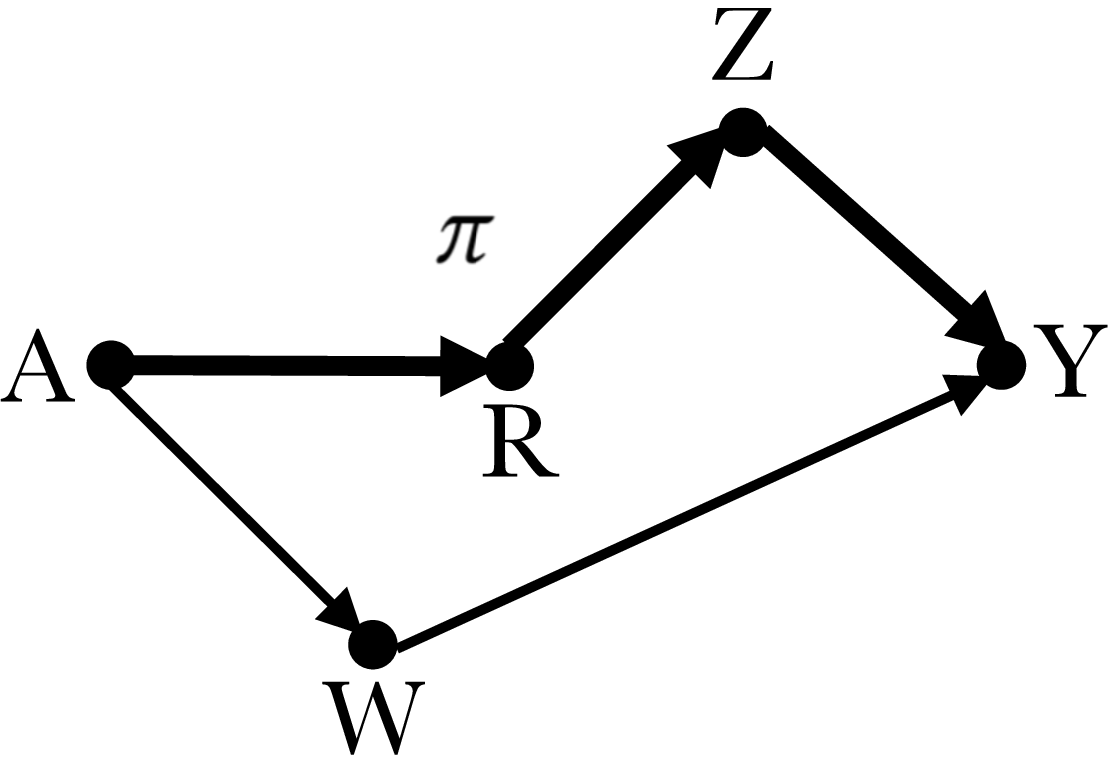}  & \makecell[l]{$\sum_{r,w,z} P(w|\;a_0)\;P(r|\;a_1)\;P(z|\;r)$\\   $P(y|\;z,w)$}\\
        \hline
      \makecell{G40\\\\\\\\} & \includegraphics [scale=0.2]{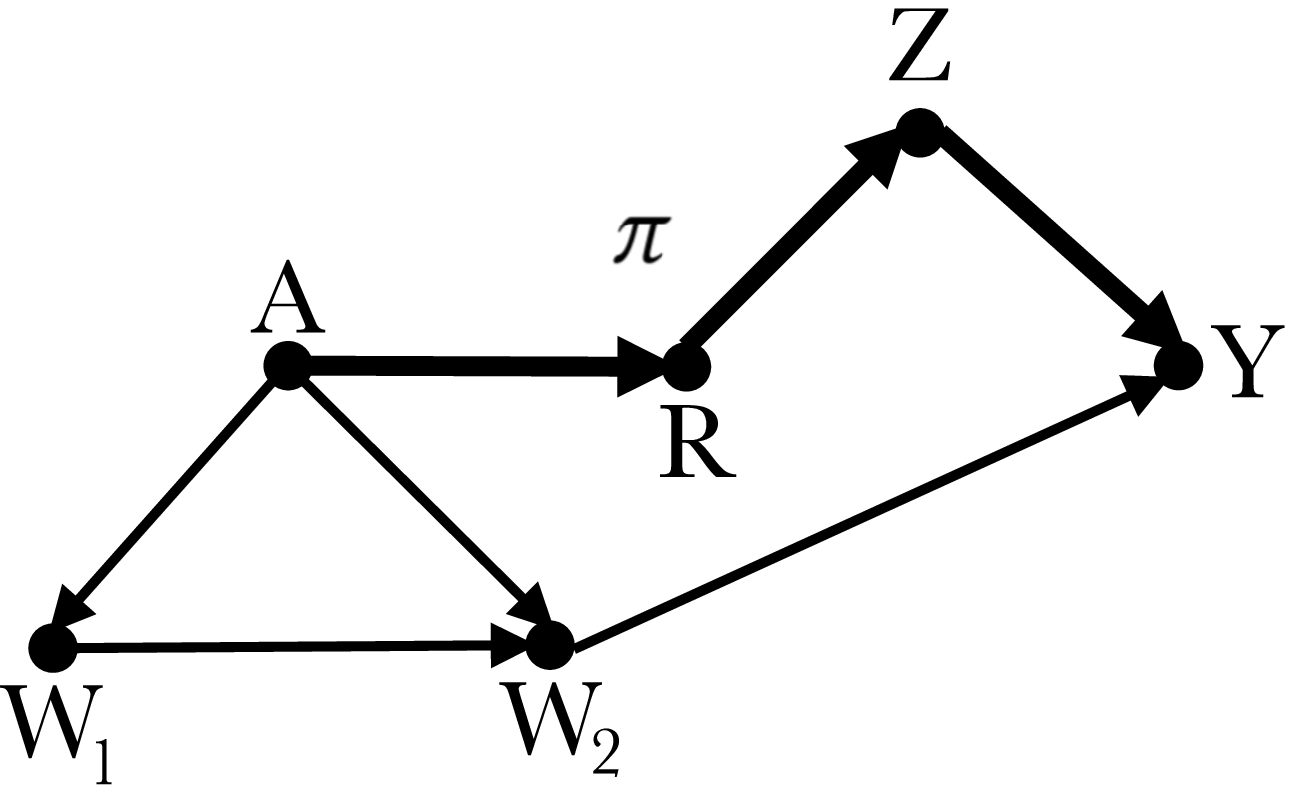}  & \makecell[l]{$ \sum_{w_1,w_2,r,z} P(w_1|\;a_0)\;P(w_2|\;a_0)\;$\\$P(r|\;a_1) P(z|\;r)\;P(y|\;z,w_2)$\\\\\\}\\
        \hline
    \end{tabular}
    \caption{Examples of Markovian graphs where PSE is identifiable due to the absence of the recanting witness.}
    \label{tab:pse}
\end{table}

Shpitser~\cite{shpitser2013counterfactual} extended the recanting witness criterion to deal with semi-Markovian models. Thus, the identifiablity of $PSE^{\pi}_{a_1,a_0}(y)$ depends on the existence (or absence) of a pattern in the graph called the recanting district. In other words, if the recanting district exists, $PSE^{\pi}_{a_1,a_0}(y)$ is not identifiable and cannot be measured from observational data. 

Given a graph $G$ and two sets of nodes: $\mathbf{A}$ and $\mathbf{Y}$ in G. Let $\pi$ be a path in $G$ starting with a node in $\mathbf{A}$ and ending with a node in $\mathbf{Y}$. Let $\mathbf{V}$ be the set of nodes not in $\mathbf{A}$ which are ancestral of $\mathbf{Y}$ via a directed path which does not intersect $\mathbf{A}$. Then a district $D$\footnote{A district is merely a c-component (introduced in Section~\ref{subsec:identsemiMarkov}).} in $G$ is called a recanting district for the $\pi$-specific effect of $A$ on $Y$ if there exist nodes $v_i,v_j \in D$, $a_i \in \mathbf{A}$, and $y_i,y_j \in \mathbf{Y}$ such that there is a path: $a_i \rightarrow v_i \cdots y_i$ in $\pi$ and an another path $a_i \rightarrow v_j \cdots y_j$ not in $\pi$. 

\begin{figure}[!h]
\vspace{-3mm}
    \subfigure []  {%
    {\includegraphics [scale=0.25]{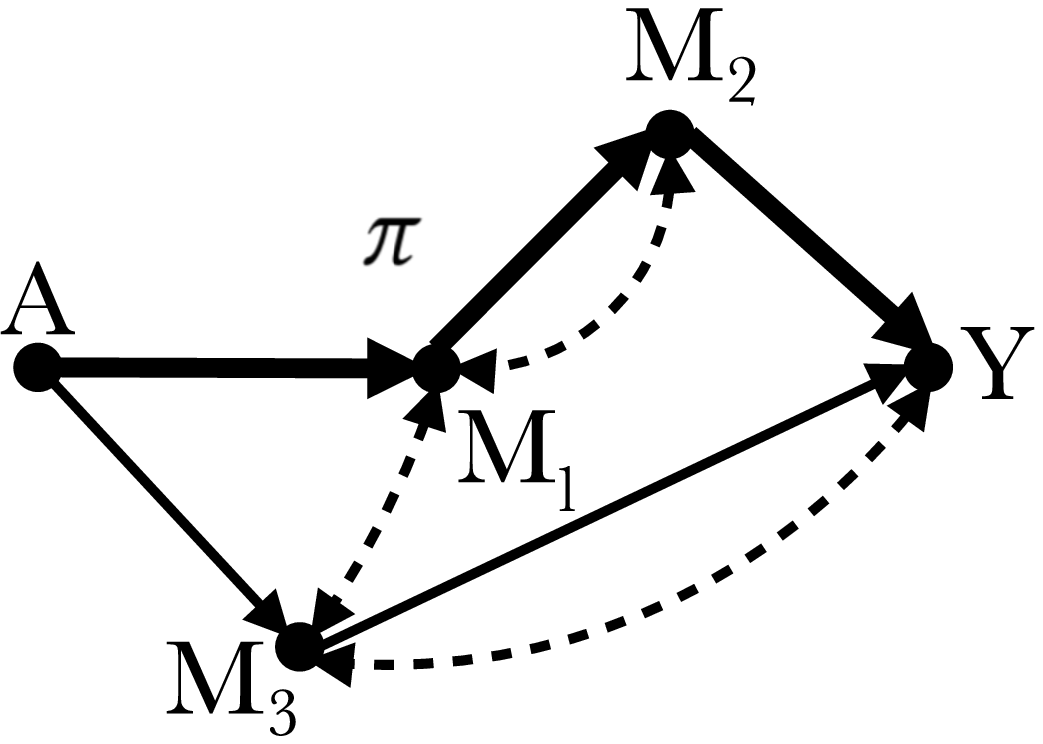} }
    \label{subfig:recanting_a}}
    \qquad 
    \subfigure [] {%
    {\includegraphics[scale=0.25]{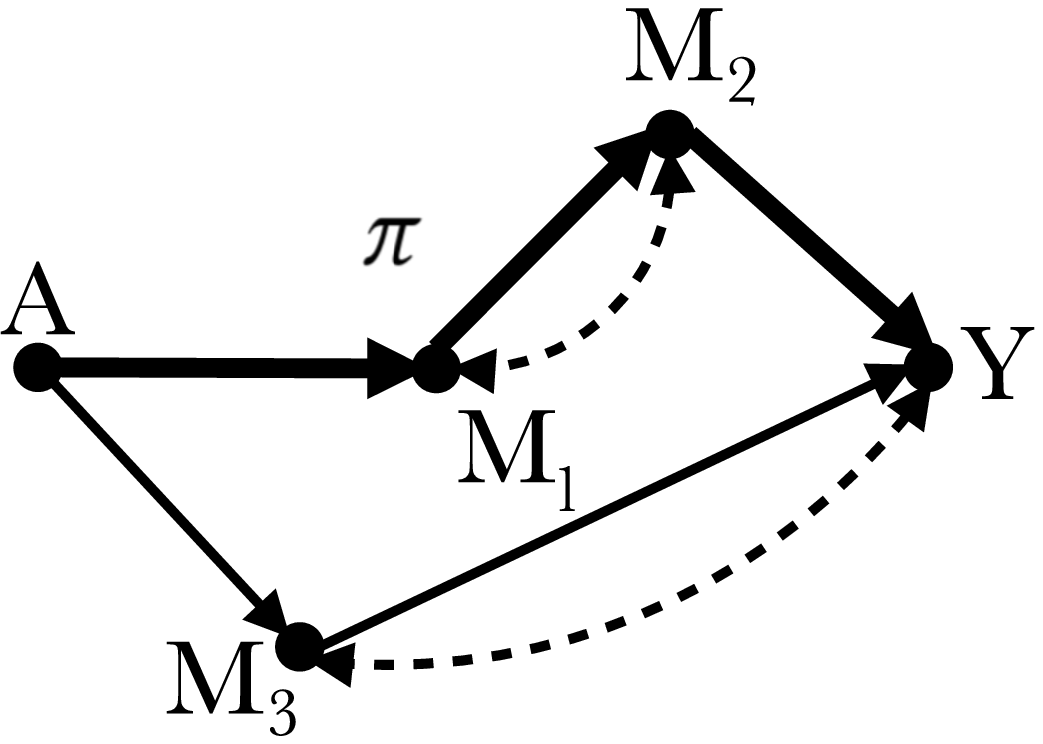} }
    \label{subfig:recanting_b}}
    \caption{The recanting district criterion satisfied. The heavy line illustrates the $\pi$ path of interest. The recanting district for Figure~\ref{subfig:recanting_a} is $\{M_1,M_2,M_3,Y\}$ while the recanting district of Figure~\ref{subfig:recanting_b} is $\{M_3\}$}
    \label{fig:recanting}
    \vspace{-3mm}
\end{figure}

As an example, consider the two semi-Markovian graphs of Figure~\ref{fig:recanting}. In either of these graphs, \textit{the recanting district} exists. In Figure~\ref{subfig:recanting_a}, the district $\{M_1,M_2,M_3,Y\}$ is recanting since the path $A \rightarrow M_1 \rightarrow M_2 \rightarrow Y$ is the $\pi$ path of interest while there exists another path not in $\pi$ connecting $A$ and $Y$, namely, $A \rightarrow M_3 \rightarrow Y$, and $M_1$ and $M_3$ belong to the same district: $\{M_1,M_2,M_3,Y\}$. Similarly, in Figure~\ref{subfig:recanting_b}, the district $\{M_3\}$ is recanting since there exist two paths connecting $A$ and $Y$: the first is $\pi$ itself ($A \rightarrow M_1 \rightarrow M_2 \rightarrow Y$) and the second is not in $\pi$ ($A \rightarrow M_3 \rightarrow Y$), and $M_3$ is its own district.

Now, if the recanting district criterion is not satisfied, the $\pi$- specific effect is identifiable and $PSE^{\pi}_{a_1,a_0}(y)$ can be calculated from observational data as follows (Theorem3~\cite{shpitser2013counterfactual}):
\begin{equation}
    \label{eq:district}
	\sum_{\substack{\mathbf{V}\backslash\{Y\}}} \prod_{D} P(D = d|\;do (E_D = e_D))
\end{equation}

where $V$ is the set of nodes not
in $A$ which are ancestral of $Y$ via a directed path which does not intersect $A$. $D$ ranges over all districts in the graph $G_V$. $E_D$ refers to nodes with directed arrows
pointing into $D$ but which are themselves not in $D$, and value assignments $d$ and $e_D$ are
assigned as follows: if any element $a$ in $A$ occurs in $E_D$ in a term $P(D = d|\;do (E_D = e_D))$,
then it is assigned a baseline value if the arrows from $a$ to elements in $D$ are not in $\pi$, and
an active value if the arrows from $a$ to elements in $D$ are in $\pi$\footnote{We use the term ``active'' to designate the value assigned to the sensitive attribute $A$ along the causal paths we are interested in and the term ``baseline'' to designate the value assigned to $A$ along all the other causal paths}. Let the baseline and the active values are $a_0$ and $a_1$, respectively. All other elements in $E_D$
and $D$ are assigned values consistent with the values indexed in the summation. 

Table~\ref{tab:recanting} shows examples of semi-Markovian models where the $\pi$-specific effect is identifiable. Note that for all these graphs, it is never the case that both an arrow in $\pi$ and an arrow not in $\pi$ emanating from the node $A$ to nodes in the same district such that these nodes are ancestors of $Y$. This implies there is no recanting district for the effect of $A$ on $Y$ hence, the $\pi$-specific effect is indeed identifiable. For instance, $G41$ involves a single district: $\{M_1,M_2,M_3,Y\}$. Applying Eq.~\ref{eq:district} yields to:
\[
 \sum_{m_1,m_2,m_3} P(m_1,m_2,m_3,y|\;do (a_1))
\]
Laveraging the general theory of identification of interventional probabilities, the above expression can be transformes as follows:
\[
    \sum_{m_1,m_2,m_3} P(m_1|\;a_1,m_3)\;P(m_2|\;m_1)\;P(m_3|\;m_1,y)\;P(y|\;m_3,m_2)
\]
$G43$, at the other hand, presents a more sophisticated case including three districts, namely, $\{M_1\}$, $\{M_2\}$, and $\{M_3,M_4,Y\}$. Thus,

\begin{align}
    &\sum_{m_1,m_2,m_3,m_4} P(m_1|\;do (a_1,m_3))\;P(m_3,m_4,Y|\;do (a_0,m_1,m_2)) \nonumber\\
    &\qquad \qquad \qquad \qquad  P(m_2|\;do (m_1,m_4)) \nonumber
\end{align}

\begin{table*}[!h]
    \begin{tabular}{|c|c|l|}
    \hline
      &Causal graph   &  \makecell{$P(y_{a_1 |_\pi, a_0 |_{\overline{\pi}}})$} \\
    \hline
       \makecell{G41\\\\\\\\}& \includegraphics [scale=0.2]{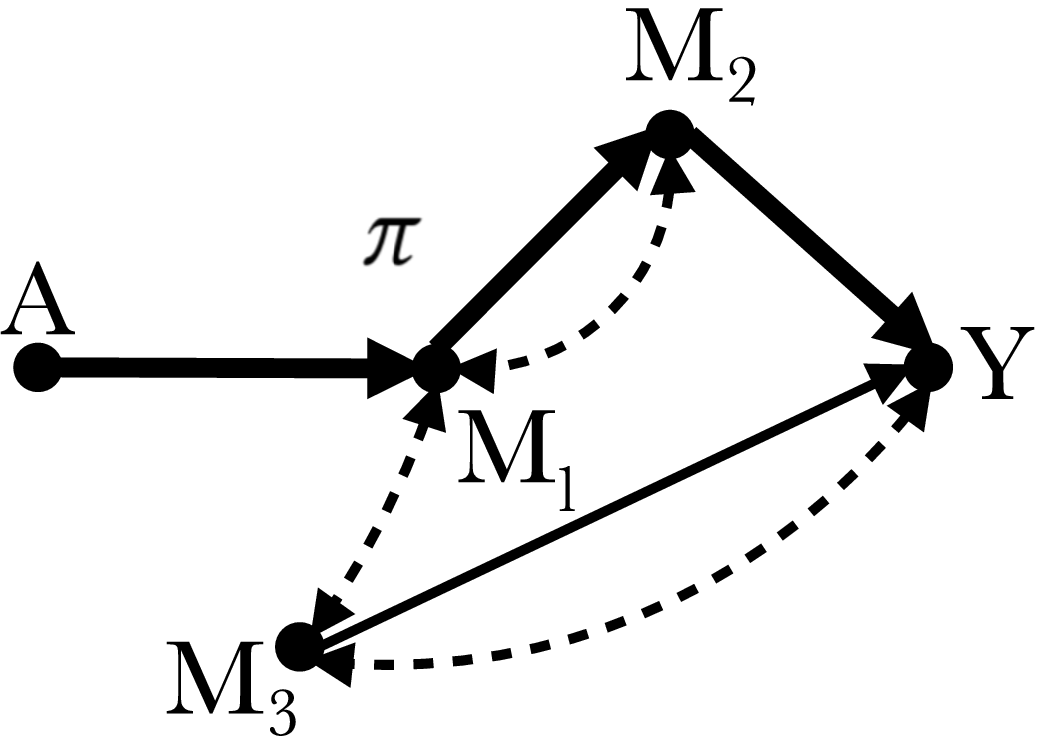} & \makecell[l]{$ \sum_{m_1,m_2,m_3} P(m_1|\;a_1,m_3)\;P(m_2|\;m_1)\;P(m_3|\;m_1,y)\;P(y|\;m_3,m_2)$\\\\\\} \\
    \hline
      \makecell{G42\\\\\\\\} & \includegraphics [scale=0.2]{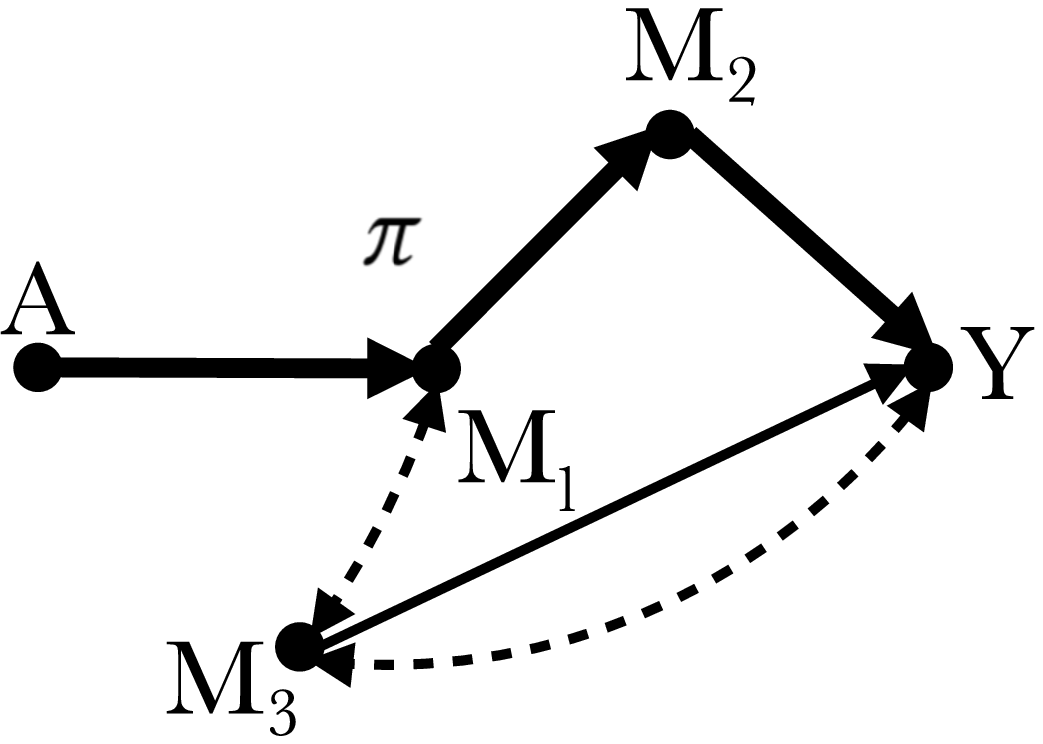}  & \makecell[l]{$ \sum_{m_1,m_2,m_3} P(m_1|\;a_1,m_3)\;P(m_3|\;m_1,y)\;P(y|\;m_3,m_2)\;P(m_2|\;m_1)$\\\\\\} \\
        \hline
      \makecell{G43\\\\\\\\} & \includegraphics [scale=0.2]{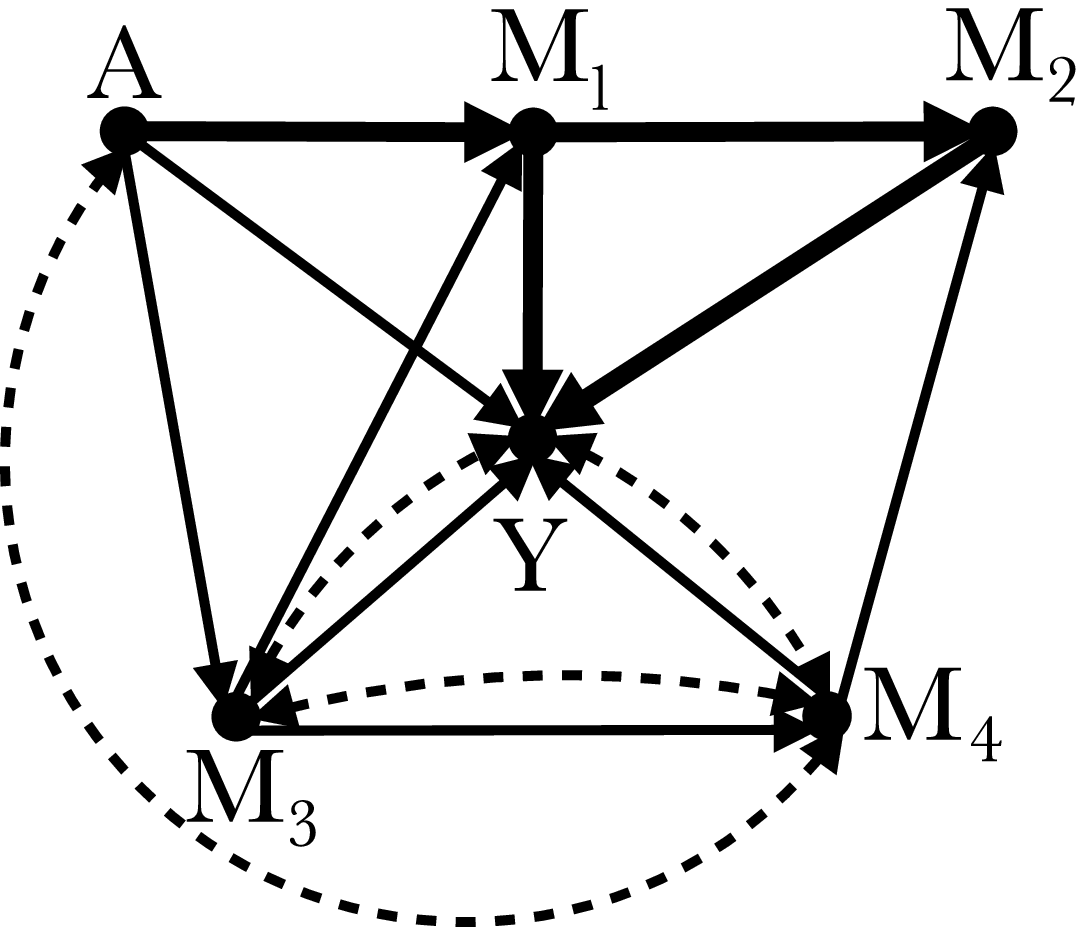}  & \makecell[l]{$ \sum_{m_1,m_2,m_3,m_4} P(m_1|\;a_1,m_3)\;\;P(m_2|\;m_1,m_4)\;P(m_3|\;a_0)\;P(m_4|\;a_0,m_3)\;P(y|\;a_0,m_1,m_2)$\\\\\\}\\
        \hline
    \end{tabular}
    \caption{Examples of semi-Markovian graphs where PSE is identifiable due to the absence of the recanting district.}
    \label{tab:recanting}
\end{table*}




\section{Conclusion}
\label{sec:conclusion}
A typical goal of causal inference in the context of discrimination discovery is establishing the causal effect of the sensitive attribute $A$ on the outcome $Y$.  Unfortunately, this may not be possible due to the identifiability problem. This paper studied the problem of identifiability as it relates to discrimination discovery. We made use of the large-scale body of work on identifiability theory to summarize the main results found in the literature. Based on  various graphical patterns, we discussed and assessed whether the causal effect of $A$ on $Y$ is identifiable.  The main identifiability results fall into three main types, namely the causal effect (intervention), the counterfactual effect and the path-specific effect. 
Finally, we note that When identification is not possible, it may still be possible to bound causal effects. 
The development of bounds for non-identifiable quantities is called  partial identifiability.

\section{Acknowledgements}
This work was supported by the European Research Council (ERC) project HYPATIA under the European Union’s Horizon 2020 research and innovation programme. Grant agreement n. 835294.

\bibliographystyle{ACM-Reference-Format}
\bibliography{biblio}

\end{document}